\documentclass[11pt]{article}

\usepackage[T1]{fontenc}
\usepackage[utf8]{inputenc}
\usepackage[english]{babel}

\usepackage[top=2cm, bottom=2cm, left=3cm, right=3cm]{geometry}

\usepackage{microtype}

\usepackage{amsmath, amssymb}

\usepackage{booktabs, tabularx, makecell} 
\usepackage{siunitx} 
\usepackage{graphicx}
\usepackage[section]{placeins} 

\usepackage{csquotes}
\usepackage{framed}

\usepackage[dvipsnames]{xcolor}

\usepackage{natbib}
\usepackage[colorlinks=true, allcolors=blue]{hyperref}
\usepackage{url}
\renewcommand{\harvardurl}[1]{\href{#1}{\nolinkurl{#1}}}

\title{Elite Polarization in European Parliamentary Speeches: a Novel Measurement Approach Using Large Language Models}

\author{Gennadii Iakovlev\thanks{%
ORCID: 0000-0002-8355-6914 \quad
Email: \href{mailto:gennadii.iakovlev@eui.eu}{gennadii.iakovlev@eui.eu}.
I am grateful to Dr.\ Francesca Chiarvesio (postdoctoral researcher in political science) for annotating the Italian validation sample.}}
\begin{document}
\maketitle

\begin{abstract}
    Theories of democratic stability, populism, and party-system crisis often point to a form of polarization that comparative research rarely measures directly: hostile relations among political elites. Existing comparative measures capture adjacent phenomena, including mass affective polarization, or elite ideological distance, but not directed mutual elite evaluation. This paper introduces the Elite Polarization Score, a measurement of out-party evaluations in parliamentary speech. Large Language Models identify political actors mentioned in parliamentary debates, recover speaker-target pairs, estimate the sentiment directed at each actor, standardize heterogeneous references into party dyads, and aggregate these evaluations into party- and parliament-level measures of mutual out-party negativity. The validity of the approach is demonstrated on parliamentary corpora from the United Kingdom, Hungary, and Italy, covering up to four decades of debate. The resulting measure is conceptually distinct from mass affective polarization, elite ideological polarization, incivility, negative campaigning, and general sentiment. Evidence from the UK case study shows that it is also empirically distinct from mass affective polarization, elite ideological polarization, and incivility. Extreme negative evaluations can also be used to locate pernicious polarization rhetoric. Validation across three countries finds no false discoveries, sentiment estimates accurate to roughly 10 percent of the scale range, and AI sensitivity that meets or exceeds that of human coders in two of three settings. Because the algorithm is multilingual, requires no task-specific training, and can be aggregated by party and quarter, it provides a scalable basis for future cross-national research on what produces elite polarization and what elite polarization itself produces.

\end{abstract}

\section{Introduction}

Mutual emotional hostility among politicians—those who write the rules rather than merely obey them—can reshape democracy far more quickly than mass‐level grievances.  Because elite attitudes are insulated only by a thin layer of institutions, not by party heuristics or media frames, spikes of animosity at the apex of power can transmit directly into legislative obstruction, constitutional hardball, and even coups. Indeed, democratic‐stability scholarship shows that emotional antipathy among governing elites almost always intensifies in the run-up to democratic breakdowns \citep{linz1978, bermeo2016, levitsky2018, luhrmann2021, wunsch2023}.

Political polarization has become a cornerstone of research on party politics and regime change.  Scholars argue that excessively large or, at the other extreme, vanishingly small—ideological divides coupled with heightened partisan affect can undermine virtually every aspect of democratic practice \citep{iyengar2012, mccoy2018, iyengar2019, druckman2020, reiljan2020, orhan2022, schedler2023, turkenburg2023}. Yet one phenomenon distinct from mass affect has remained understudied: the mutual hostility inside political elites \citep{hetherington2001, banda2018, iyengar2019, druckman2020}.

The demand for reliable measures of any form of polarization within political science has grown in recent years, but the available tools remain limited. Traditional methods, such as aggregating survey data, are typically confined to specific countries and time periods, typically once in five years. Likewise, sentiment or contextual analyses often require extensive training and manual coding, which restricts most projects to a single country \citep{abercrombie2020}. This methodological gap had thwarted cross-country studies until recent advances in artificial intelligence.

This article proposes a novel methodology for measuring elite polarization using Large Language Models (LLMs). The key conceptual advance is disaggregating the unit of analysis from the speech to the individual reference: the LLM identifies every instance where one politician mentions another, gauges the speaker's evaluative attitude toward that referent, and distinguishes evaluations of the actor from assessments of the actor's circumstances. Without actor-subject detection, speech-level sentiment analysis can at most measure the average temperature of a policy debate, not the specific relationships between democratic adversaries. Beyond this, LLMs require no task-specific training and operate across languages without additional adaptation, making a cross-national, multilingual panel dataset tractable for the first time \citep{kheiri2023, mens2023, abercrombie2020}.

I introduce a multi-stage algorithm that submits standardized parliamentary speeches one by one to an LLM via API query. The model autonomously identifies every instance where politicians mention one another, evaluates the speaker's sentiment toward the referenced counterpart using contextual cues rather than keyword matches, and records both the speaker and target name. Downstream algorithms then classify the resulting references into party dyads and aggregate hundreds of thousands of evaluations into a time-series Elite Polarization index. A customized validation protocol assesses the recall of actor mentions and the inter-annotator agreement in sentiment scoring.

Certain single steps in this actor and subject detection approach could be executed using traditional NLP techniques. For example, dictionary-based or machine-learning Named Entity Recognition (NER) methods \citep{li2020} might capture some references to political entities. However, these techniques would likely only detect explicit mentions of governments and political parties but not their MPs, and their effectiveness would be limited to a few large, predominantly English-speaking democracies. Next, one could approximate the speaker's attitude toward these actors by applying dictionary-based sentiment analysis to the words surrounding the entity names. Without the benefit of contextual embedding, traditional methods would not be able to distinguish whether these words are used to actually characterise the speaker's stance towards a referred entity. Words proximate to an entity name frequently describe that entity's circumstances rather than the speaker's attitude toward it, and may not refer to the entity at all. Hence, both steps of extracting a subject and extracting the evaluation of a subject would suffer from extremely low precision and recall, with errors compounding over the process (see, among many others, \citealt{caselli2015s}).

To assess the performance and face validity of the approach, I apply it to three European countries: Hungary, the United Kingdom, and Italy. The selection is based on a range of party-system types, electoral rules, and democratic trajectories. Hungary combines documented high mass affective polarization \citep{orhan2022}, a collapsed two-party system \citep{powelltucker2014}, and a trajectory of democratic deterioration throughout the entire time sample, only later interrupted in spring 2026; its non-Indo-European language also constitutes a stress test for the cross-lingual claims of the methodology. The United Kingdom offers a contrasting case of low mass polarization \citep{orhan2022}, a stable  two and a half party system, and no backsliding; its extensive NLP literature \citep{goet2019} allows external validation of the index. Italy represents an intermediate case: moderate mass polarization, a proportional system with high fragmentation, repeated coalition turnover, and no backsliding \citep{chiaramonte2017}.  Methodologically, the approach builds on recent work showing that LLMs can be used to position political texts on substantive dimensions through direct prompting \citep{gallego2025}, extended here to reference-level sentiment with explicit actor resolution.

    This paper is organized as follows. In the next section, I review the concept of polarization, comparing findings from the regime change and party politics scholarships, and discussing how these perspectives problematize polarization in the democratic process. The theory section grounds the Elite Polarization construct in, and states the scope conditions under which the score serves as a proxy for elite affective polarization, distinguishes the measure from neighbouring constructs, and presents the scoring formula with its two departures from Reiljan's \citeyearpar{reiljan2020} template. Section 3 outlines the involvement of natural language processing (NLP) in political science, especially in parliamentary speech. The methodology section explains the LLM-based pipeline, the validation protocol: model benchmarking, adjudicated by LLM-assisted fuzzy matching. Section 6 presents vignette case studies on polarization in three countries: Hungary from 2002 to 2023, the United Kingdom from 1987 to 2020, and Italy from 2013 to 2022. Finally, Section 7 concludes and discusses implications for future research.

\section{State of the Art}

In the last decade, the concept of polarization has gained prominence in both regime change and party politics research. Scholarship on regime change typically adopts a country-level perspective, focusing on political elites through mid- to small-N designs (often country-level case studies). By contrast, party politics research has focused on individuals, employing mostly large-N individual-level designs—such as mass surveys and experiments—guided by well-established methodologies in political psychology and sociology.

A substantial portion of the literature on the current democratic crisis examines incremental processes that can lead to the collapse of party systems or even democratic backsliding (see, for example, \citealt{linz1978, bermeo2016, luhrmann2021, wunsch2023}). In societies that experience democratic backsliding, a recurring pattern emerges: before the breakdown of once-stable party systems, polarization intensifies \citep{chiaramonte2017, bertoa2021, mainwaring2021}. Ruling elites abandon "mutual forbearance" in their competition \citep{levitsky2018} and exhibit antidemocratic tendencies, such as publicly prioritizing their own victory over institutional norms and threatening opponents and journalists. According to this view, elite affective polarization undermines democracy when it escalates into "hatred," leading parties to endorse "abusive, illegal, and even authoritarian acts" \citep{levitsky2018}. In other words, when politicians lose "democratic trust" in their adversaries and begin to view them as an existential threat \citep{schedler2023}, a democratic regime is endangered.

From a more ideology-focused perspective, polarization becomes harmful when it empowers radical "anti-system movements" \citep{haggard2021, luhrmann2021}. Whether the distance between factions is emotional or ideological, these theories suggest that in excessively polarized systems, the importance of democratic institutions diminishes, paving the way for anti-democratic measures. Anticipating similar behavior from their rivals, would-be autocrats engage in what Nancy Bermeo calls "executive aggrandizement": curtailing media freedom, controlling the judiciary, eliminating opponents, and amending electoral rules \citep{bermeo2016, wunsch2023}. Empirical evidence generally supports this intuition, although the only study demonstrating a robust link between polarization and democratic backsliding (based on aggregated survey data) is limited to around one hundred country-years \citep{orhan2022}.

Within party systems scholarship, polarization is recognized as more multifaceted. Some degree of both elite and mass polarization is seen as integral to the healthy functioning of democracy: moderate elite ideological polarization can stimulate stronger citizen polarization \citep{zingher2018, druckman2020}, which in turn can increase political engagement and improve the quality of political opinions \citep{hetherington2001, abramowitz2008, druckman2013}. However, excessive polarization harms democratic quality. Heightened elite ideological polarization impedes citizens' capacity for critical reflection, reinforcing prior beliefs even when confronted with clear evidence to the contrary \citep{druckman2013}. Extreme affective polarization in masses similarly diminishes trust in democratic institutions, thereby undermining their legitimacy \citep{fiorina2008, iyengar2012, mccoy2018, iyengar2019, reiljan2020, orhan2022, schedler2023}. Notably, cross-national evidence shows that the steep rise in affective polarization documented for the United States is far from universal: AP trends vary substantially across democracies, with several European countries exhibiting stable or even declining levels \citep{boxell2024}.

Despite advances in measuring mass affective polarization, these methods have rarely been applied to the study of elites. First, methodologically sophisticated party politics research has generally treated affect as a mass-level phenomenon, analyzing elites primarily in terms of ideological or policy positions \citep{hetherington2001, banda2018, iyengar2019, druckman2020}. Second, as discussed below, cross-country sentiment analysis of elite speeches was previously unfeasible—and almost non-existent—before the advent of LLMs. Finally, most polarization research relies heavily on US data and seldom addresses multiparty systems or autocratizing contexts \citep{boxell2024, wagner2021}.

Because of these gaps, political scientists have lacked the means to conduct an uninterrupted, cross-national longitudinal investigation of elite polarization. As a result, there remains "little to no research identifying the \textit{mechanisms} underlying polarization" \citep[p.~133]{iyengar2019}. In other words, we still know relatively little about the conditions under which elite polarization grows, stabilizes, or declines. One possibility is that elite polarization reflects a relatively durable feature of a party system, rooted in a pre-existing "perception that opponents are untrustworthy" \citep{iyengar2019, schedler2023} and evolving slowly in response to structural factors. Another possibility is that elite polarization is more reactive: once grievances are articulated by political actors, hostility can accumulate rapidly, but may also recede once the triggering conflict loses salience. Without comparable time-series data, however, it has been difficult to adjudicate between these hypotheses.

The same measurement gap has also prevented scholars from treating elite polarization as an explanatory variable in its own right. This is especially important because many processes central to democratic stability depend on elite agency. The literature on party-system crisis and populism, for example, debates whether populism originates primarily from political supply or from popular demand \citep{pappas2019, moffitt2016}. Yet even when popular demand is strong, it is elites who decide whether to adopt populist practices \citep{guiso2024}: antagonizing the political nation against established elites, using bait-and-switch tactics, making undeliverable promises, or pursuing harmful and senseless reforms \citep{guiso2024}. More broadly, while existing research suggests that excessive elite hostility is detrimental to democratic quality, we still lack systematic evidence on how much elite polarization is required before it begins to destabilize democratic competition. A longitudinal measure of elite polarization would therefore allow scholars to examine whether it anticipates populist escalation, institutional hardball, and democratic backsliding.

\section{Theory}

\subsection{Defining Polarization}

In political science scholarship, the term ``polarization'' has gained wide currency, and publications on the topic continue to grow rapidly \citep{rollicke2023, schedler2023}. This expansion has made the concept increasingly ambiguous, much like what occurred with the term ``populism'' a decade ago. Polarization may refer to ideological distance, affective hostility, mass- or elite-level conflict, dyadic or multiparty competition, and attitudes or behaviors. This section clarifies these distinctions and defines elite polarization as MPs' directed evaluations of domestic out-parties in parliamentary speech.

Substantively, polarization is usually divided into \textit{ideological} and \textit{affective} dimensions. Ideological polarization involves differences in policy preferences and ideological stances \citep{dimaggio1996, dewilde2019, norris2019}, especially when these opinions cluster at opposite ends of an ideological spectrum. In this sense, ideological polarization implies movement away from a moderate center toward the extremes in the overall distribution of opinions \citep{fiorina2008}. Affective polarization, by contrast, refers to emotional divisions, negative evaluations, or even hostility between political groups \citep{iyengar2019, reiljan2020}. Most scholars understand affective polarization as the distance between in-party liking and out-party dislike, although some approaches focus on out-party dislike alone \citep{goldin2026, gidron2023, levitsky2018}.
Polarization also varies according to the actors among whom it is studied. As Table \ref{tab:typology} shows, it can be examined among \textit{elites} or among the \textit{mass public}. In practice, however, the resulting two-by-two typology has mostly produced two dimensions instead of four. Ideological polarization is most often treated as an elite-level phenomenon, whereas affective polarization is usually studied at the mass level, to the extent that the terms ``affective polarization'' and ``mass polarization'' are sometimes used interchangeably. Measurement strategies reflect this division. Studies of ideological polarization commonly rely on \textit{party manifestos} \citep{dimaggio1996, dewilde2019}, including large-scale projects such as ParlGov and the Manifesto Project. Less frequently, ideological polarization is examined through NLP analyses of parliamentary debates, mostly in the US and the UK \citep{goet2019, hopkins2022}. Affective polarization, by contrast, is usually measured through mass surveys. The object of this affect varies across studies \citep{rollicke2023}: some examine evaluations of ordinary supporters of other parties \citep{iyengar2019}, while others focus on evaluations of politicians from other parties \citep{reiljan2020}. There are also attempts to measure citizens' ideological stances \citep{norris2019}, as well as studies of affective polarization among political elites using surveys in Spain \citep{sanchez2024}, Canada \citep{lucas2025}, and the US \citep{enders2021}.
    
    \begin{table}[h]
        \centering
        \caption{Polarization: Four Analytical Corners}
        \label{tab:typology}
        \begin{tabular}{lcc}
            \toprule
            & \textbf{Masses} & \textbf{Elites} \\
            \midrule
            \textbf{Ideology} & Issue distance between partisans & \textit{Ideological distance between parties} \\
            \textbf{Affect} & \textit{Partisan in-group like \& out-group dislike} & \textbf{Elite} in-party like \& \textbf{out-party dislike }\\
            \bottomrule
        \end{tabular}
    \end{table}
    
A further ambiguity concerns what scholars mean by ``affect.'' In practice, the term covers not only immediate and unprocessed emotions, but also more stable evaluations formed over time \citep[p.~349]{schedler2023}. The most influential definition by \citet{iyengar2019}, for example, treats affective polarization as the combination of in-party favoritism and out-party animosity, which are attitudes. Some studies extend the concept further to behavioral outcomes, such as inter-party marriage patterns \citep{iyengar2012}. In this paper, I use the narrower language of \textit{evaluations} rather than emotions, because parliamentary speech captures public, processed, and institutionally filtered expressions rather than immediate feelings.

Finally, polarization varies in the number of poles it assumes. Although the term itself suggests two poles, the expected structure depends heavily on context. Scholars working on the US and UK often assume two party-poles \citep{fiorina2008, iyengar2019}, whereas scholars of European party systems consider multiple poles \citep{sartori2005, reiljan2020, orhan2022}.  In this case, the polarization is only possible as a system-level characteristic. The metaphor of poles can also be extended to one's in- and out-groups (for instance, in- and out- party)\citep{somer2025}, thus making polarization also an individual- or party- level characteristic.
    
 \subsection{Elite Polarization via Out-Party Evaluation}
    
    I operationalize \textit{elite polarization} as the aggregate evaluation of domestic out-parties in parliamentary debates. More simply, the measure captures how MPs speak about their political opponents in official parliamentary speech.\footnote{That said, before party names are attributed, my algorithm does collect cases in which foreign actors, domestic NGOs, or businesses are mentioned. These references are mechanically excluded from the index because such actors are not MPs and do not have party affiliations, but remain in the raw data.} The measure only records evaluations of identifiable domestic political actors; and it is multiparty-sensitive because evaluations are interpreted in relation to the size and relevance of the parties to which they are directed. In Table \ref{tab:typology}, it occupies the lower right analytical corner of what is understood as polarization.
    
    I refrain from calling this measure ``affective polarization'' to avoid confusion with a direct analogue of survey-based affective mass polarization for three reasons. First, I do not include in-party evaluations. Second, the measure captures processed public evaluations rather than immediate emotions. Third, it is measured using elite parliamentary speech rather than mass-level partisan feeling surveys. 
    
    The standard survey-based definition of affective polarization understands it as the distance between in-party warmth and out-party hostility \citep{iyengar2019, reiljan2020}; the empirical constraints of the parliamentary setting make the in-party part awkward. For mass survey research, the in-party component is substantively and methodologically important. Citizens may identify with a party strongly, weakly, or not at all; they may like their preferred party, or simply dislike it less than the alternatives. In-party evaluations can therefore carry information comparable to out-party evaluations. Measuring the distance between in- and out-party evaluations also helps address individual-level scale bias, since respondents may interpret feeling thermometers differently across cultures \citep[p.~193]{king2004}, while a net distance between in- and out-group evaluations mitigates this problem \citep[pp.~381--383]{reiljan2020}. Finally, mass survey respondents can express negative views of their own party without any consequences.
    
    Parliamentary speech differs in all three respects. First, MPs are not analogous to ordinary citizens: party membership, candidate selection, and legislative discipline make strong party identification a reasonable baseline assumption. Second, public criticism of one's own party is likely to be constrained by party rules, reputational concerns, and career incentives. Third, the intensity of intra-party debate may itself depend on party-system fragmentation, which is higher in two-party systems \citep[pp.~18--19]{noamziblatt2019}. Including in-party evaluations could therefore make elite polarization appear lower in two-party systems, introducing bias into cross-country comparisons.
    
    There is also an empirical reason to exclude in-party evaluations. MPs refer to their own party one-third as often as they refer to out-parties.\footnote{Out-to-in reference ratios: 3.16 (Hungary), 2.76
    (United Kingdom), 2.30 (Italy).} The remaining in-party references are also strategically censored: party rules, reputational concerns, and career incentives discourage public criticism of one's own party, so the surviving in-party valence concentrates near mild positive values. Retaining an in-party component that is smaller in volume and institutionally constrained in valence would therefore add noise without adding much substantive information beyond what out-party evaluations already capture.
    
    On individual- and party- level, the party-level component of Elite Polarization measurement closely relates to the branch of literature that preceded affective polarization---negative campaining---"the act of attacking the opponent on his programme, values, record or character instead of advocating his own" \cite[p.~11]{nai2016}. However, this tradition did not differentiate between ideological and affective difference; additionally, negative campaigning was often perceived as something extreme, approaching the definitions of perniciously polarizing speech \cite{somer2025}. Most of the literature in that predominantly US-based tradition that qualitatively analyzed the public debates \cite[p.~12-14]{nai2016}, rather than parliamentary rhetoric. Additionally, negative campaigning was never considered a system-level characteristic. The Elite Polarization Score, in some way, extends this tradition by isolating affective hostility from ideological components, and by aggregating party-level evaluations into a system-level index that is comparable across countries and over time.
    
    \subsection{What the Measure Does Not Capture}
    
    Because the concept of polarization has become exceptionally crowded, it is worth specifying the boundaries of the Elite Polarization Score. The EPS captures directed evaluations of domestic out-parties in parliamentary speech. It does not capture all forms of ideological distance, incivility, or mass-level partisan hostility.
    
    First, the EPS is not an ideology scale. It does not measure the distance between parties on a left-right or policy dimension; that function belongs to manifesto-based or roll-call-based scaling approaches \citep{dimaggio1996, dewilde2019}. The EPS captures how MPs \textit{evaluate} their opponents in public, which can diverge substantially from their positional distance. Two parties may be ideologically close yet rhetorically hostile, or ideologically distant yet rhetorically cordial. That said, the UK case shows a moderate correlation ($r = 0.5$) between the EPS and an established ideological polarization index, consistent with the affective--ideological relationship reported in mass survey research \citep{reiljan2020}.
    
    Second, the EPS is not a measure of the general sentiment of parliamentary debate. General sentiment analysis of plenary speeches aggregates every evaluative or even emotional expression in a speech, including assessments of policy outcomes, international events, and domestic conditions. The EPS records only evaluations that are explicitly directed at identifiable domestic political actors. For instance, a speech condemning economic conditions is not polarizing under this measure; a speech denouncing the government for those conditions is.
    
    Third, the EPS is not an index of incivility or pernicious polarization. Incivility research focuses on norm violations, such as personal attacks, ridicule, or name-calling, which are only a subset of negative evaluations \citep{somer2025}. The EPS captures the full range of out-party negative evaluations, from mild criticism ($-1$) to hostility ($-5$), and includes positive evaluations as well. The distribution of extreme negative values can serve as a proxy for searching for instances of pernicious polarization, but the index itself is not designed to flag specific rhetorical violations.

\subsubsection{Limitations}
    
 The main limitation of the EPS is that parliamentary speech is not a transparent record of MPs' private attitudes. Speeches are public, strategic, institutionally regulated, and shaped by parliamentary procedure. Compared to media or social media discourse, parliamentary debates allow for more standardized cross-national comparison. Nevertheless, they provide a skewed sample of elite discourse. The EPS excludes non-parliamentary political figures and less formal, often more emotional, forms of public speech, such as media interviews, campaign speeches, and social media posts. It should therefore be interpreted as a measure of elite hostility that enters the official parliamentary record, not as a measure of the full universe of elite political communication. The measure, therefore, captures performed out-party evaluations rather than unfiltered affect. In this sense, it sits at the outer boundary of the affective polarization literature: it measures processed, public evaluations rather than immediate emotional reactions. If this limitation is accepted, the EPS can be understood as an elite affective polarization score, bearing in mind that it does not capture the difference between in- and out- groups.
    
This does not make the measure empirically irrelevant to the extreme end of affective elite polarization -- pernicious polarization. Studies of political elites often focus less on the balance between in- and out-group evaluations and more on extreme forms of out-group hostility: the perception that opponents are untrustworthy \citep{iyengar2012}, democratic enemies \citep{schedler2023}, or existential threats \citep{levitsky2018}. As I show in Appendix Table~A3, more than a quarter of the references with the most negative evaluations delegitimize their opponents, which virtually never occurs among moderately negative evaluations. Thus, even though \textit{pernicious elite polarization} \citep{somer2025} appears only at the negative extreme of the EPS scale, those extreme values can be useful as its proxy.
    
\subsection{Elite Polarization Score Formulae}

To create the Elite Polarization Score (EPS), I build on the conceptual foundations of Reiljan's \citeyearpar{reiljan2020} measure of mass affective polarization in multi-party settings. Reiljan weights each evaluation on both sides by vote share: by the vote share of the respondent's party and by the vote share of the party being evaluated. A negative evaluation of a 10\% party therefore contributes one-fifth as much to system-level polarization as the same evaluation of a 50\% party. \citet{wagner2021} similarly shows that in multiparty systems, the number and size of parties substantially shape both the level and the dynamics of affective polarization in ways that two-party frameworks cannot capture.

The measure developed here adopts the same size-weighting logic, with two adaptations to parliamentary speech. First, as discussed above, I exclude in-party sentiment because MPs identify with their own parties and rarely refer to them in parliamentary speeches. Second, while Reiljan weights both sides by vote share, parliamentary speech provides a more direct signal of which out-parties a speaker is actually engaging with: the frequency with which they reference each one. I therefore retain vote share at the speaker (parliament) level, where it captures electoral weight, but use reference share at the target level, where it captures observed parliamentary attention.

In words, the measure can be summarized as:
\textbf{\[
\text{Elite Polarization Score}_{t}
=
-
\text{vote-share-weighted mean of party-level out-party evaluations}_{t}.
\]}
The party-level out-party evaluations that enter this system-level average are
themselves attention-weighted. For a given speaker party, I weight evaluations of out-parties by the proportion of parliamentary references each out-party receives in the period under analysis. This follows from the structure of the data. In survey-based measures of mass affective polarization, party-size weights correct for unequal probabilities of selection: supporters of some parties may be over- or under-represented in survey samples relative to the population. Parliamentary speech does not require the same survey-style correction, because the denominator is not a sample of citizens but the observed population of references in the parliamentary corpus. The reference share $\mathrm{Ref.\,share}_{m,t}$ therefore captures the share of parliamentary attention that party $m$ commands in period $t$.

The system-level score answers a different question. At the party level, the measure asks: when party $n$ speaks about its out-parties, how negative is the mix of opponents it actually discusses? At the parliament level, the question is: how negative is the party system as a whole? Some parties may attract many references without having equivalent influence over the party system---or vice versa. I therefore aggregate party-level out-party evaluations using each party's size and electoral vote share.

The resulting procedure has two stages. First, for each speaker party $n$, I estimate its average evaluation of all out-parties, weighting each out-party by the parliamentary attention it receives. Second, I aggregate these party-level out-party evaluations to the parliament level, weighting parties by vote share and reversing the sign so that greater negativity corresponds to higher polarization.\footnote{To recap, high elite polarization is the predominance of
negative evaluations.} Data permitting, the same logic can be applied at coalition, party, and even individual levels at different temporal granularities. Below, I present the quarterly party and parliament-level version.

\begin{itemize}
    \item $\mathrm{Like}_{n \to m,t}$: the mean quarterly sentiment of references
    made by MPs from party $n$ toward out-party $m$ in period $t$, on the
    $[-5,+5]$ scale extracted by the LLM.
    \item $\mathrm{Ref.\,share}_{m,t}$: party $m$'s share of all parliamentary
    references in period $t$.
    \item $\mathrm{Vote\,share}_{n,t}$: party $n$'s electoral weight in period
    $t$, carried forward between elections.
\end{itemize}

Party $n$'s average out-party evaluation is the weighted mean of its evaluations of all other parties, where each out-party's weight is its reference share normalized among parties other than $n$:
\begin{equation}
    \mathrm{Like}^{\mathrm{out}}_{n,t} \;=\;
    \sum_{\substack{m=1 \\ m \neq n}}^{N}
    \left[
        \mathrm{Like}_{n \to m,t} \cdot
        \frac{\mathrm{Ref.\,share}_{m,t}}{1 - \mathrm{Ref.\,share}_{n,t}}
    \right].
    \label{eq:party_out_eval}
\end{equation}
The denominator $1 - \mathrm{Ref.\,share}_{n,t}$ is the total reference share of all parties other than $n$. This normalization ensures that the out-party weights for each speaker party sum to one.

The parliament-level EPS then aggregates these party-level out-party evaluations using party vote shares, negating the result so that predominance of negative evaluations yields a positive polarization score:
\begin{equation}
    \mathrm{EPS}_{\text{parliament},\,t} \;=\;
    -\sum_{n=1}^{N}
    \left(
        \mathrm{Like}^{\mathrm{out}}_{n,t} \cdot
        \mathrm{Vote\,share}_{n,t}
    \right).
    \label{eq:eps}
\end{equation}

Positive values of $\mathrm{Like}^{\mathrm{out}}_{n,t}$ indicate warmer average
evaluations of out-parties by party $n$, while negative values indicate more
critical evaluations. The EPS reverses this sign, so higher values correspond to greater average out-party negativity in parliament, and therefore higher elite polarization. 

Appendix provides analogues of Figure~\ref{fig:same_scale_main} for alternative weighting configurations: reference share at both party and parliamentary levels, and vote share at both levels. The resulting patterns are broadly similar. However, weighting by vote share alone creates sharp changes at election points, which do not happen in the observed discourse, whereas weighting by reference share at both levels is theoretically more vulnerable to disproportionality when a party receives a disproportionately large or small share of parliamentary attention. Additionally, the Appendix provides a variant with in-group evaluations, which is also substantively very close.

The resulting number is a weighted out-party sentiment score on the same $[-5,+5]$ scale as the underlying reference-level evaluations, with the sign reversed at the parliament level. A score of $0$ indicates neutrality toward out-parties. Negative EPS values mean that MPs, on average, speak of opponents positively; positive EPS values mean that MPs, on average, speak of opponents negatively. The theoretical bounds of the index are therefore $-5$ and $+5$, but these extremes are not empirically plausible: a score of $-5$ would require
praise for every out-party at every mention, while a score of $+5$ would require uniform hostility. In practice, given the large number of unavoidable neutral contextual references, values approaching $3$ already indicate an extremely polarized parliamentary system.

\section{Natural Language Processing of Parliamentary Speech}
This section overviews the development of natural language processing, focusing particularly on sentiment analysis techniques over the past three decades. It describes the evolution from rule-based or dictionary analysis to supervised machine learning approaches, culminating in neural networks and deep learning, and the current trend of using pre-trained transformer neural networks, that is, LLMs. In doing so, it illustrates their applications in political science.

The most straightforward way to implement text classification, particularly sentiment analysis, is to create a dictionary where specific words represent certain categories, such as positive or negative emotions. However, this approach is inherently limited in accuracy and portability. The main problem is that dictionary analysis overlooks context, irony, or even negations, which often completely alter the meaning of a phrase; thus, accuracy scores rarely exceed 65\% for one-step tasks, much simpler than the one presented in this article. Second, dictionaries cannot be easily transferred between different contexts without significant adaptation, limiting studies to a single setting. Finally, these approaches often require expertise from computational linguists.

Hopkins and Schickler used a dictionary-based approach to analyze the language in American state party platforms over a century \citep{hopkins2022}. By constructing specialized dictionaries of Republican and Democrat political terms and phrases, they measure the degree of ideological elite polarization and nationalization in parliamentary discourse. However, the century they model had to be divided into three periods, each with its own adopted dictionary. They observed that polarization levels were relatively low in the mid-20th century but began increasing significantly from the 1980s onward.

A more recent and complicated application of rule-based analysis measures populist discourse in speeches delivered in the European Parliament from 1999 to 2014 \citep{hunger2024}. Just as this paper, Hunger's work introduces a two-step dictionary approach: the first step distinguishes mentions of 'people' and 'elites'; the second targets morally charged descriptions. In this application, the accuracy score reaches 75\%.

The rule-based epoch was gradually superseded by supervised and unsupervised machine-learning techniques. Instead of relying on hand-coded rules, these methods enable computers to learn patterns from existing corpora of training data. Machine learning approaches range from simpler methods, such as bag-of-words models, to complex deep learning algorithms used in LLMs. The application workflow differs: in the rule-based approach case, a researcher manually creates a dictionary; in the supervised machine learning, a researcher manually trains a model from scratch; with LLMs, a researcher applies an out-of-the-box pre-trained model, adjusting it to their needs via prompt engineering or, at most, fine-tuning.

Compared to dictionary analysis, machine learning can achieve very high accuracy scores, reaching 80–90\% similarity with human coders. However, these techniques require significantly more time to train for a specific purpose and are often even more context-dependent than rule-based approaches. Extensive training would normally make similar projects dominated by computer science rather than political science scholars \citep{abercrombie2020}. In addition, such studies almost never took more than a couple of countries \citep{abercrombie2020}.

Several supervised and unsupervised machine learning studies have analyzed ideological polarization in UK parliamentary debates over two centuries \citep{goet2019}. In their case, the accuracy was approximately 60\% for pre-WWII data and around 70–80\% after WWII. A supervised machine learning method can assess more theoretically sophisticated and ambiguous topics. Di Cocco and Monechi  \citeyear{dicocco2022} measure the degree of populism in European political party manifestos across six countries over the last two decades. This approach achieved an F1 score of 88\% for classification against human coding and a 70–80\% correlation with other indices. Supervised machine learning models can also perform more complex tasks, such as automatically identifying and classifying protest events from news articles across different countries and contexts \citep{hurriyetoglu2021}. By training their algorithm on annotated datasets, the latter study achieved impressive F1 scores that, when translated to accuracy, are around 85\%.

LLMs differ from the supervised and fine-tuned approaches described above in that they are pre-trained by their developers on vast, heterogeneous text corpora and then adapted to specific tasks through prompt engineering rather than retraining \citep{juravsky2026, bommasani2021}. Applied to sentiment and stance annotation, they match or outperform trained crowd workers on standard political-text tasks \citep{kheiri2023, mens2023, gilardi2023}. Unlike earlier neural architectures, they transfer across languages without additional adaptation \citep{lai2024}---a property that makes multilingual comparative datasets feasible. Despite this technical maturity, cross-national longitudinal applications that use LLMs to construct new theory-relevant measures---rather than to replicate existing indices---remain rare in the published literature.

A further development now on the horizon is the emergence of \textit{agentic} AI systems, in which LLMs autonomously orchestrate multi-step workflows---retrieving corpora, dispatching annotation sub-tasks in parallel, aggregating results, and triggering quality checks---without continuous human supervision \citep{wangsurvey2024}. The pipeline presented in this paper remains largely sequential and requires a human in the loop, but the architecture is compatible with full agentic execution. Running the annotation pipeline in agentic mode across the ParlaMint corpora for all twenty-nine covered EU member states and the European Parliament \citep{erjavec2023} would, in principle, produce a twenty-year EU-wide elite polarization panel--a scale of data collection that was structurally impossible before this generation of models.
\section{Analysis Procedure}
\subsection{Algorithms}
I developed an automated text-analysis procedure that uses a large language model to identify and evaluate references to political actors in parliamentary speeches. Implemented through a custom set of R functions, the algorithm submits each speech to the model together with a standardized prompt, and contextual country and year. The model returns structured information on the actors mentioned, the direction of the reference, and the evaluative tone of the statement. This prompt is used in the procedure:
 \begin{framed}
You are a political scientist analyzing a parliamentary speech delivered in Hungary in 2016\footnote{The example shows the Hungarian version of the prompt; country and year fields are substituted automatically for each speech.} by a politician. Provide only a CSV table. Follow these steps:

\medskip
\textbf{Identification of Political Actors:}\\
1.1. List all politicians, political parties, and other political entities mentioned in the speech.\\
1.2. Exclude non-political entities from the list, such as competitions, general groups (like businessmen), and sectors (like local transport).

\medskip
\textbf{Detailed Analysis of References to Political Actors:}\\
2.1. For each identified political actor in step 1.1, describe the specific sections of the speech where they are mentioned. Provide a detailed description of the context in which each political actor is mentioned, focusing on how the speaker refers to them and the circumstances surrounding these references.\\
2.2. For each identified political actor in step 1.1, provide a rationale for considering the mentioned actor as political, and not excluding them during step 1.2.

\medskip
\textbf{Analysis of Sentiment Toward Political Actors:}\\
3.1. Locate the sections of the speech that refer to the identified political actors from step 1.\\
3.2. Analyze the speaker's emotional attitude toward each political actor. Focus on the speaker's sentiment about the actors themselves, not about their situations or conditions. Use a scale from –5 (strongly dislikes) to +5 (strongly likes), with 0 standing for neutrality.
\end{framed}

Then follows the plain, untokenized text of the speech. The attention layers of LLMs made the quality of the results significantly better if the text was analyzed in its original language instead of the machine-translated version, even in the case of non-Indo-European Hungarian language. The results were then compiled into a comprehensive table. 

The next stage standardizes the heterogeneous list of extracted political actors. It first classifies each reference into one of four categories: party or party member, government, institution, and foreign institution. It then uses the country and year of the speech to associate named politicians, offices, titles, and spelling variants with the relevant party. Many surface forms can refer to the same actor. For instance, in a Hungarian speech from 2011, entries such as `Prime Minister', `Orbán', or `miniszterelnök' all point to Viktor Orbán, who belonged to the then-governing Fidesz party; these references are therefore recoded as `Fidesz'\footnote{This lookup task is not included in the main extraction prompt because it requires extensive contextual knowledge about politicians, offices, party affiliations, and country-specific naming conventions. Separating extraction from standardization makes the procedure more efficient: many references can be resolved through deterministic lookup tables, while the remaining ambiguous entities are deduplicated and queried only once per unique entity--country--year combination, rather than at every occurrence. This reduces the number of LLM-assisted lookup queries by orders of magnitude and keeps the process manageable for human supervision.}

The standardization stage combines deterministic and LLM-assisted procedures. Country-specific lookup scripts search for known word patterns, offices, party labels, and spelling variants, while machine-learning NER packages such as spaCy help identify named entities that require disambiguation. Together, these procedures resolve roughly half of the references extracted in the previous pass. The remaining unique unresolved entities are then passed to an LLM with the following prompt:
     \begin{framed}
\medskip
Identify the political party affiliations of the individuals mentioned in the Hungarian parliamentary speeches, as of the speech year indicated in separate columns.\\
   Determine the political party for each named individual.\\
  For entries that are individuals, list their name followed by the political party
  they are associated with.\\
  For other entities\\
  1. If a political party is mentioned, simply replicate the line; \\
  2. For government bodies, label them as 'government';\\
  3. For non-governmental institutions, like NGOs, label them as 'institution';\\
  5. For foreign institutions, use 'foreign institution'.\\
   Ensure that each individual's name is paired with their party name.
 \end{framed}
The resulting lookup table is then manually checked. This human review is necessary because the number of frequently used unique entities runs into the thousands, while only a subset of them refer to MPs or domestic political parties and therefore enter the EPS. Government bodies, institutions, foreign actors, and other non-party entities are retained in the raw extraction output but excluded from the party-based polarization index. Finally, a separate aggregation script combines the resulting speaker--target--sentiment dyads into the time-series Elite Polarization Score using the formula described above.

The performance of the algorithm depends critically on prompt design. I developed the current prompt through iterative refinement: early versions produced unreliable entity detection until the task was restructured as a six-step chain-of-thought. The two most consequential advances were (1) separating actor identification (steps 1.1--1.2) from sentiment evaluation (step 3), which prevents the model from conflating detection with assessment; and (2) adding an explicit exclusion criterion (step 1.2) to filter geographic objects, social groups, and procedural forms of address. Requesting a rationale for each evaluation (step 2.1) substantially improved both the replicability and the coherence of sentiment assignments\footnote{The prompt is incompatible with older or smaller models such as ChatGPT 3.5, LLaMA 3 8B, or numerous BERT derivatives known to me, which lack the context length and instruction-following capability required by the chain-of-thought structure.} Providing country and year as context in the system prompt further sharpened the model's use of domain-relevant knowledge.\footnote{The speaker’s name, affiliation, and dates could increase extraction quality even more, but they could bias the LLM’s responses to attribute antipathy to out-groups.}

Sentiment classification is performed in a purely zero-shot manner---relying on the LLM's internal representations rather than providing labelled exemplars. I do not supply example outputs for each point on the $-5$ to $+5$ scale, nor do I define the emotion labels explicitly. The prompt already employs chain-of-thought walkthrough, that is, it involves recurrent passes asking an LLM to check the results of previous steps. Hence, inserting few-shot exemplars at the classification step would complicate and obscure the reasoning flow.\footnote{Which, as of spring 2026, would make the prompt incompatible with the models below 240B} This approach targets the model's latent encoding of affective states; examples of the built-in sentiment mappings that emerge under this prompt appear in the Appendix.

LLMs from the GPT-4 generation onward are capable of navigating the nuances of political texts beyond the reach of earlier machine learning techniques.\footnote{Including open-weight competitors of the same generation, such as LLaMA 3.1 70B and Mistral Small 22B v24.09.} They can differentiate a speaker's attitude toward an actor from the actor's circumstance---when an entity faces hardship, the overall tone is negative, but the speaker's evaluation of that entity may be neutral or even sympathetic. They can also articulate their reasoning in natural language, which fosters falsifiability. Sentiment agreement with human coders was very high in most models released from the second half of 2023 onward, but attributing the direction of a sentiment was a bottleneck until the GPT-4 generation: weaker models frequently confused subject and agent, attributing ``suffocating the cities'' to the ``opposition'' in the phrase ``the ruling party suffocates the opposition-governed cities.'' By 2026, leading open-weight models have largely resolved this, as confirmed by the validation results in Table~\ref{tab:validation}.

\subsection{Quality Control Procedures}

The validation protocol assesses two properties of the annotation pipeline. The first is \textit{sensitivity}, or recall: what fraction of all politically relevant actor mentions in a speech does the AI identify. The second is \textit{sentiment agreement}, i.e.\ how closely the AI's ordinal evaluation of each matched reference corresponds to that of a trained human coder, measured as mean absolute error (MAE) on the $-5$ to $+5$ scale.

All proprietary models were queried between mid-March and late April 2025, using fixed API endpoints and pinned open-weight versions. The production model is GPT-4o-2024-08-06 for Hungary and Italy, and GPT-4o-mini-2024-07-18 for the UK long-run series; open-weight alternatives are Gemma~4  (26B, MoE, 4B active parameters), MiniMax M2.7 (230B, MoE, 10B active parameters); and DeepSeek-V3.2 (685B parameters) used for fuzzy-matching.

The conventional approach to sensitivity validation treats the human annotation as the ground-truth ceiling and measures what fraction of it the AI recovers. This is appropriate when the human annotator is the more exhaustive coder. That assumption does not hold in the present setting. Starting from the GPT-4 generation, LLMs tend to identify more actor references per speech than a human coder working under realistic time constraints \citep{gilardi2023}. Treating human coding as the denominator penalises the AI for \textit{valid} references the human missed, misclassifying genuine coverage as a false discovery. For instance, in Hungary, the AI identifies roughly 39\% more unique references per speech than the human coder, exceeding the human count in 65\% of speeches.

To evaluate annotators on equal terms, I construct a \textit{supergold} union standard via LLM-assisted fuzzy matching. A human coder annotates a random sample of non-consecutive speeches using the same prompt text provided to the AI, and the AI processes the same speeches. Any actor identified by the human is included in the supergold. Actors identified only by the AI are then validated against the source speech and retained only if they correspond to genuine textual mentions. The resulting supergold therefore, consists of all verifiably valid references identified by either annotator. A human manually overviews the resulting union, correcting the AI-assisted merge and checking the validity of all findings. Then, both the AI and the human are scored against this joint ceiling.

The reconciliation of heterogeneous name forms is a politically informed entity-linking task. For example, references such as ``Orbán'', ``Prime Minister'', ``miniszterelnök'', and ``Viktor Orbán'' may all point to the same actor in a given country-year context. At the scale of this project, resolving these references manually is infeasible, while purely rule-based string-matching methods are poorly suited to titles, abbreviations, inflected forms, and context-dependent references. I therefore use DeepSeek-V3.2 to propose candidate matches, with final adjudication performed by a human coder.

Table~\ref{tab:validation} reports a false-discovery rate of zero across all country--model combinations in the validation samples. This means that every AI-identified reference in test samples had a genuine textual anchor in the source speech. For this reason, I am not using F1 score, and report recall alone: the precision component would have no variation. 
That being said, estimate concerns textual false discoveries, not whether every identified actor belongs in the final polarization index. The extraction prompt is deliberately broad. The LLM is instructed to identify political actors and political entities, including governments, institutions, foreign actors, and social groups. This broader extraction strategy increases recall and makes the dataset reusable for analyses beyond the party-based polarization index. All of those references are excluded mechanically from the EPS because they are not domestic party actors, but they are retained in the raw data

Table~\ref{tab:validation} reports the full set of results. Sentiment agreement is quantified as MAE between matched AI and human ordinal scores. MAE of 0.85 (Hungary) and 1.48 (UK) on a $\pm5$ scale, comparable to inter-annotator agreement in manual sentiment-coding studies of political speech \citep{abercrombie2020}.  The scatterplots comparing human and AI sentiment annotation are reported in the Appendix. The direction of the annotator asymmetry differs across the three countries.

In Hungary the AI covers 86.3\% of the supergold union against the human coder's 62.3\% (ratio\,=\,1.39), making the model the more exhaustive annotator; in the United Kingdom the human outperforms the AI (80.0\% vs 65.0\%; ratio\,=\,0.81); Italy shows a modest human advantage (64.9\% vs 58.6\%; ratio\,=\,0.90), compounded by the annotation-format asymmetry described in the table note.

Along with the GPT-4o that was used in production as this article was being created, I benchmark SoTA open-weight LLMs for both methodological and practical reasons. Methodologically, this choice follows a growing literature that treats openness as an advantage for reproducibility, auditability, and long-run reuse. As this article was being finalized, smaller and mid-sized open models had become sufficient production tools for text classification. More than that, as Table~\ref{tab:validation} shows, the size of the model does not matter for the extraction quality, with 26 and 230 billion parameters models showing roughly similar performance. This makes the proposed method potentially scalable for the entire scope of countries with available corpora.

\begin{table*}[t]
\caption{Performance of LLMs for political actor detection and sentiment analysis\textsuperscript{c}}
\label{tab:validation}
\centering
\scriptsize
\setlength{\tabcolsep}{5pt}
\renewcommand{\arraystretch}{1.15}
\begin{tabular}{@{}l c r r r r r r c@{}}
\toprule
\textbf{Model} & \textbf{CC} &
\makecell{\textbf{N refs}\\(\textbf{spch.})} &
\makecell{\textbf{Sens\textsubscript{AI}}\\(\%)} &
\makecell{\textbf{Sens\textsubscript{hum}}\\(\%)} &
\makecell{\textbf{AI/hum}\\ratio} &
\makecell{\textbf{Pearson}\\$r$ (sent.)} &
\makecell{\textbf{MAE}\\(sent.)} &
\textbf{FDR} \\
\midrule
\multicolumn{9}{@{}l}{\textit{Full validation}} \\[1pt]
GPT-4o & HU & 556 (92) & 86.3 & 62.3 & 1.39 & 0.699 & 0.85 & 0 \\
GPT-4o & IT & 147 (45) & 58.6 & 64.9 & 0.90 & 0.707 & 1.53 & 0 \\
GPT-4o-mini & UK & 683 (51) & 65.0 & 80.0 & 0.81 & 0.590 & 1.48 & 0 \\
\midrule
\multicolumn{9}{@{}l}{\textit{Open-weight alternatives}} \\[1pt]
Gemma 4 26B\textsuperscript{a} & HU & 571 (92) & 81.1 & 60.1 & 1.35 & 0.611 & 0.85 & 0 \\
Gemma 4 26B & IT & 143 (45) & 60.2 & 72.2 & 0.83 & 0.626 & 1.18 & 0 \\
Gemma 4 26B & UK & 655 (50) & 69.0 & 85.4 & 0.81 & 0.467 & 1.02 & 0 \\
MiniMax M2.7\textsuperscript{b} & HU & 569 (92) & 75.6 & 61.5 & 1.23 & 0.753 & 1.04 & 0 \\
MiniMax M2.7 & IT & 151 (45) & 57.9 & 68.4 & 0.85 & 0.983 & 1.32 & 0 \\
MiniMax M2.7 & UK & 708 (50) & 69.3 & 79.5 & 0.87 & 0.533 & 1.38 & 0 \\
\bottomrule
\end{tabular}

\vspace{3pt}
{\scriptsize\emph{%
CC: HU = Hungary, IT = Italy, UK = United Kingdom.
Sens\textsubscript{AI/hum}: share of the human$\cup$AI supergold union recovered by each annotator.
N refs: supergold union size; speeches in parentheses.
Pearson $r$: sentiment correlation for matched actor pairs ($-5$ to $+5$).
MAE: mean absolute error on sentiment.
FDR: false-discovery rate; all AI-only references verified present in source text.\\
\textsuperscript{a} Gemma 4 26B: MoE, 4\,B active parameters.\quad
\textsuperscript{b} MiniMax M2.7: 230B, MoE, 10\,B active parameters.\quad
\textsuperscript{c} Reference matching, supergold construction, and false-discovery adjudication performed by DeepSeek-V3.2 (${\approx}$685\,B parameters).}}
\end{table*}

\section{Vignette Case Studies}
The three case studies presented in this section pursue the methodological goal. They demonstrate the validity of the proposed concept and its measurement, illustrating that those are coherent and sometimes responsive to significant events, such as crises or elections. Simply put, this section demonstrates why the Elite Polarization Score is not a mere statistical noise. The UK case further reinforces this by comparing the metric with established measures of polarization.

\begin{figure}[htbp]
\centering
\includegraphics[width=\linewidth]{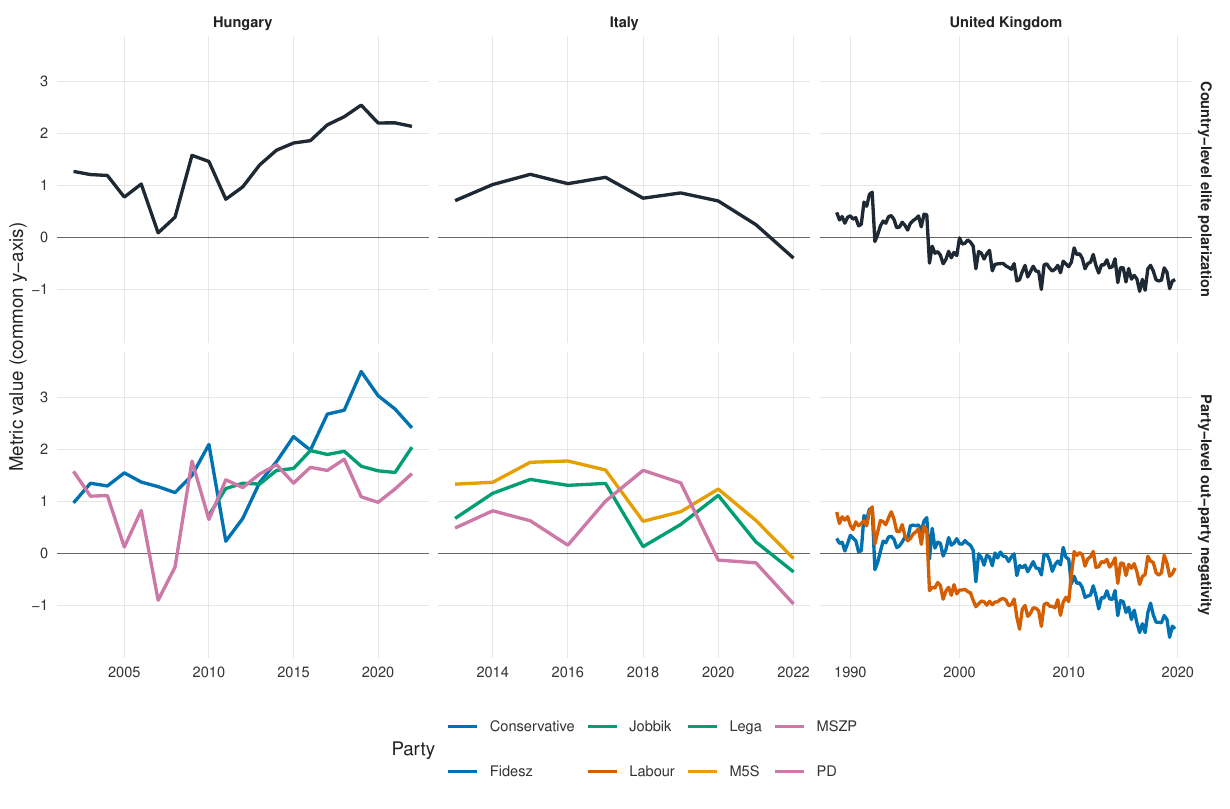}
\caption{Elite polarization and party-level out-party negativity on common scales across the three cases. The top row reports country-level elite polarization. The bottom row reports party-level out-party negativity for the parties plotted in the country narratives. All panels use common y-axis limits, so one unit occupies the same visual distance across countries; x-axis windows differ by country.\label{fig:same_scale_main}}
\end{figure}

For the case of Hungary, I use the ParlaMint dataset \citep{erjavec2023} for the years 2014--2023 and ParlLawSpeech \citep{sebokparllaw2024}---a corpus of Hungarian plenary speeches ---for the years 2002--2014. To keep the ParlaMint extension comparable with the earlier Hungarian corpus, I retain post-2020 speeches only when their XML context identifies them as interpellations or urgent questions; this corpus contains about 56,000 references to political entities.  For Italy, I use the ParlaMint dataset \citep{erjavec2023} that contains 37,500 political speeches, resulting in about 167,700 references to political entities. For the case of the UK, I use the massive ParlSpeech dataset\citep{rauh_parlspeech_2020} that includes all parliamentary speeches and covers the years 1988-2020, has 500,000 speeches, and yields about 2,460,000 political-actor references.
Among the three cases, the UK has the lowest level of elite polarization, while Hungary has the highest, aligning with the mass-level affective polarization scores reported by \cite{orhan2022}. The Figure~\ref{fig:same_scale_main} plots the elite polarization and party-level out-party negativity on common scales across the three cases. To reduce visual noize, the Hungarian and Italian data is aggregated by year, while higher definition UK data is aggregated by quarters, which makes the line oscilate more. 

\FloatBarrier

\subsection{Elite Polarization in Hungary in the years 2002-2023}

Since the end of the Cold War, Hungary has maintained a de facto two-party system in which the communist successor Hungarian Socialist Party (MSZP) and the center-right Fidesz alternated in power. In the closely contested 2002 and 2006 elections, the Socialists won by a narrow margin. However, in the aftermath of the 2006 elections, Hungary experienced a major political crisis. A post-electoral inter-party speech delivered by then-Prime Minister Ferenc Gyurcsány, in which he used obscene language and admitted that his party had deliberately misled the public about the nation's financial health to secure electoral victory, triggered a major political crisis \citep{enyedi2016}. This led to the most violent street protests since 1956, during which the protesters built barricades and captured a T-34 tank to drive several meters towards the riot police.  Later, the police dispersed the protesters with force, with hundreds wounded \citep{pribersky2008}.

In the long run, the 2006 events, followed by the 2008 financial crisis are often associated with an overall loss of MSZP's credibility \citep{meyer-sahling2020}, its landslide electoral defeat in 2010, and eventual decay after 2014. Another important change that followed this crisis is the drastic rise of the then far-right party Jobbik, which has gained popularity since then and would become an important parliamentary party after the 2010 elections \citep{bernhard2021}. In the parliamentary debates, however, these events have resulted only in a slight rise in Fidesz's out-party negativity before the 2006 elections.

In the 2010 elections, Fidesz won by a landslide with the degree of trust towards the previous government being as low as 20\% \citep{huber2023}. Almost unrestricted, in the following years, Viktor Orban started to change the legislation, limiting the freedom of the media in 2010, changing the constitution in 2011, gerrymandering the electoral districts, abolishing the veto powers of the constitutional court, and packing it with loyal judges \citep{greskovits2016, meyer-sahling2020}. In other words, they could relatively easily commit what \citep{bermeo2016} calls executive aggrandizement.
\begin{figure}[htbp]
\centering
\includegraphics[width=0.8\linewidth]{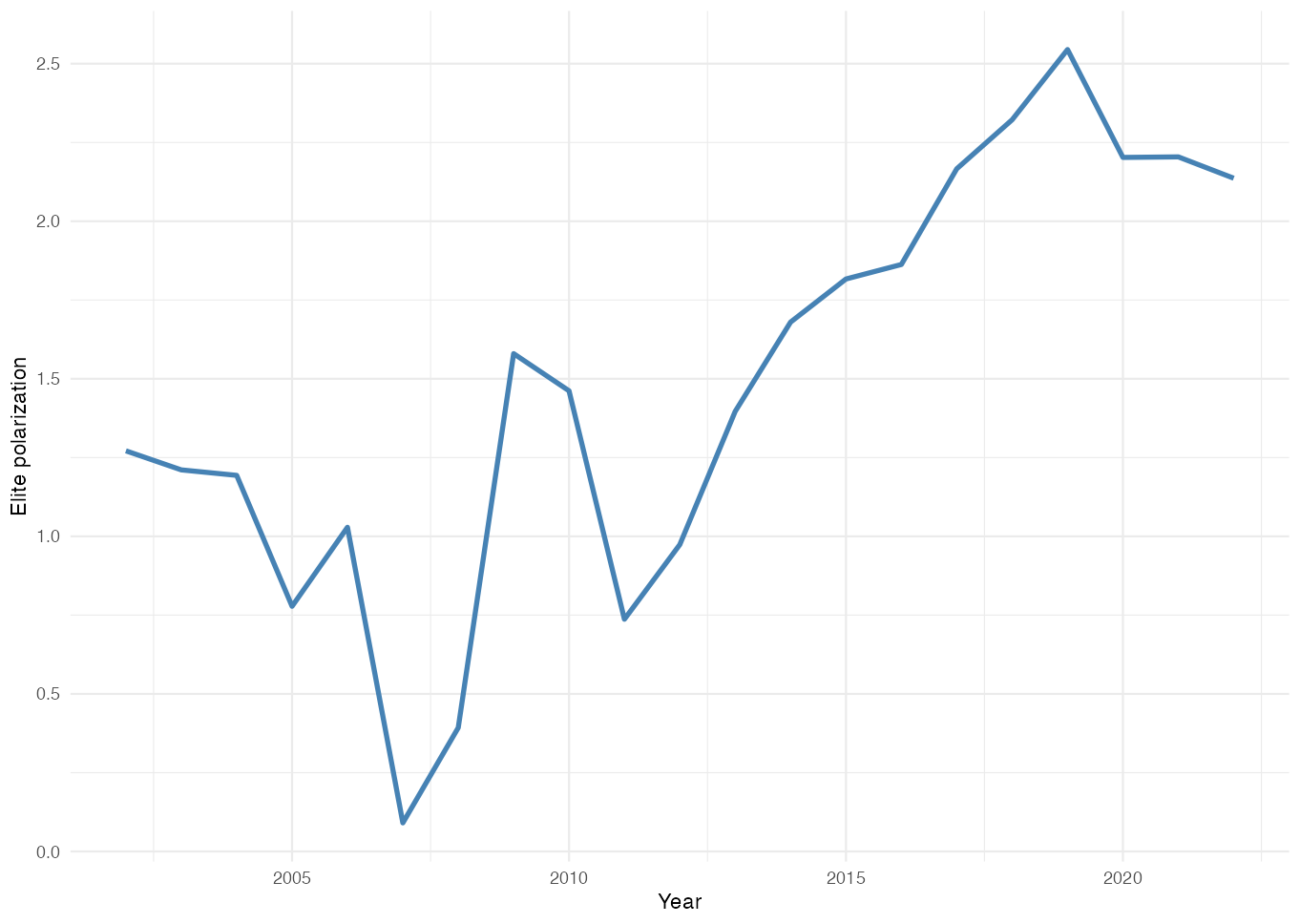}
\caption{Hungarian elite polarization, 2002--2023\label{fig:apshu}}
\end{figure}

\begin{figure}[htbp]
\centering
\includegraphics[width=0.8\linewidth]{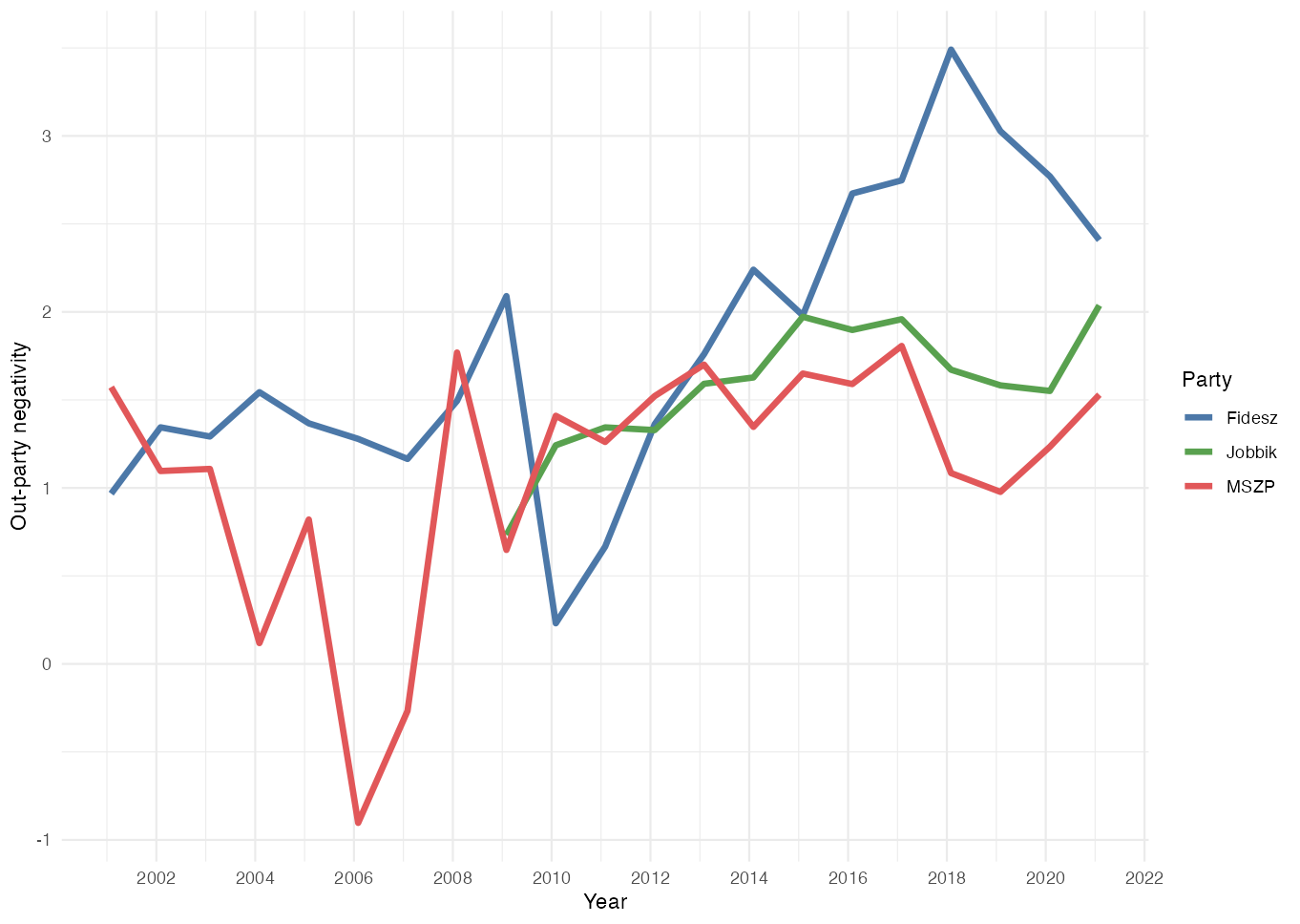}
\caption{Out-party negativity of major Hungarian parties, by year\label{fig:apshu2}}
\end{figure}

If one is to follow the theoretical expectations from the regime change scholarship \citep{levitsky2018}, one would expect elite polarization to rise before Viktor Orban came to power in 2010 and to rise further as democratic institutions were being restructured. The parliamentary series only partly follows this expectation. Figure~\ref{fig:apshu} shows a visible increase around 2009-2010, followed by a fall in 2011 and then a gradual recovery through 2014. The clearest and most sustained increase comes later, peaking in 2018--2019 rather than during the initial phase of executive aggrandizement.

This pattern reflects the party-level dynamics shown in Figure~\ref{fig:apshu2}. Once Fidesz entered government, its out-party negativity fell sharply in 2011, but it did not remain uniformly low throughout the whole 2010--2014 period; by 2014 it had returned to a high level. Because the largest party is not showing high negativity, the aggregate Elite Polarization index also remains moderate in those formative years. Opposition criticism matters, but it is filtered through the electoral weight of the parties producing it. After 2015, the party-level lines became more consistently elevated, with Jobbik rising around the migration-crisis period and Fidesz reaching its highest values in 2018--2019. The Hungarian case, therefore, suggests that elite polarization captures the consolidation and later reconfiguration of a political regime more clearly than the immediate moment of institutional takeover.

\subsection{Elite Polarization in the UK in the years 1987-2020}

The United Kingdom from 1987 to 2020 provides an example of low affective mass polarization, one of the lowest in the \cite{orhan2022}  dataset, a surprisingly stable majoritarian two-party system, and a stable democracy without any signs of backsliding. Additionally, the UK allows me to compare my measure of polarization with other NLP analyses of parliamentary speeches \citep{goet2019}. This healthy system faced several political challenges, including the shift of the leftist party to the center under New Labour in 1997, the Scottish independence referendum in 2014, and Brexit from 2016 to 2020, alongside frequent government changes. Like in Hungary, party-level out-party negativity in the UK tends to respond more to party-level events than to country-level events, affecting the aggregated country-level score accordingly.

\begin{figure}[htbp]
\centering
\includegraphics[width=0.8\linewidth]{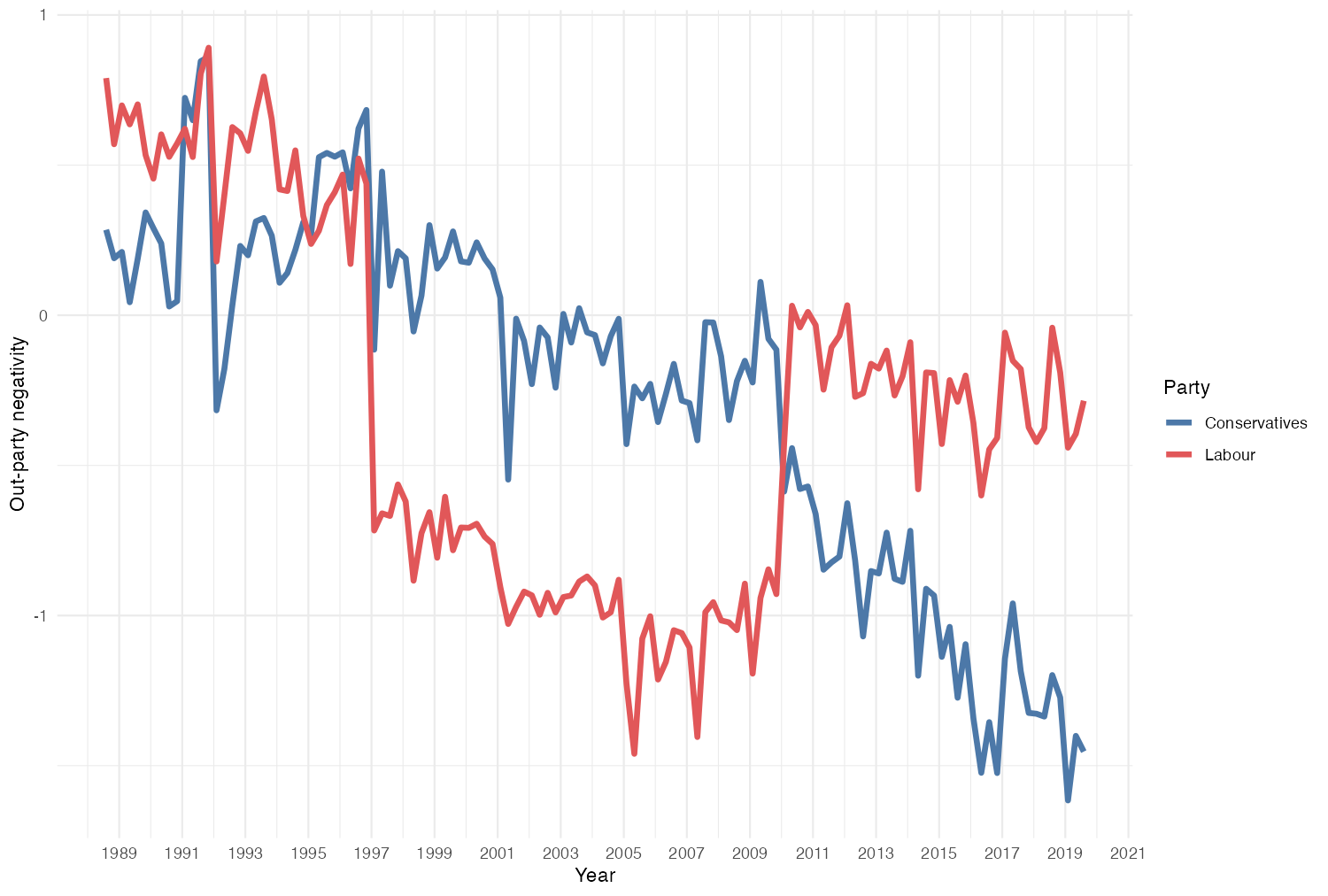}
\caption{Out-party negativity of UK Labour and Conservative parties, by quarter\label{fig:aps_uk_q}}
\end{figure}

\begin{figure}[htbp]
\centering
\includegraphics[width=0.8\linewidth]{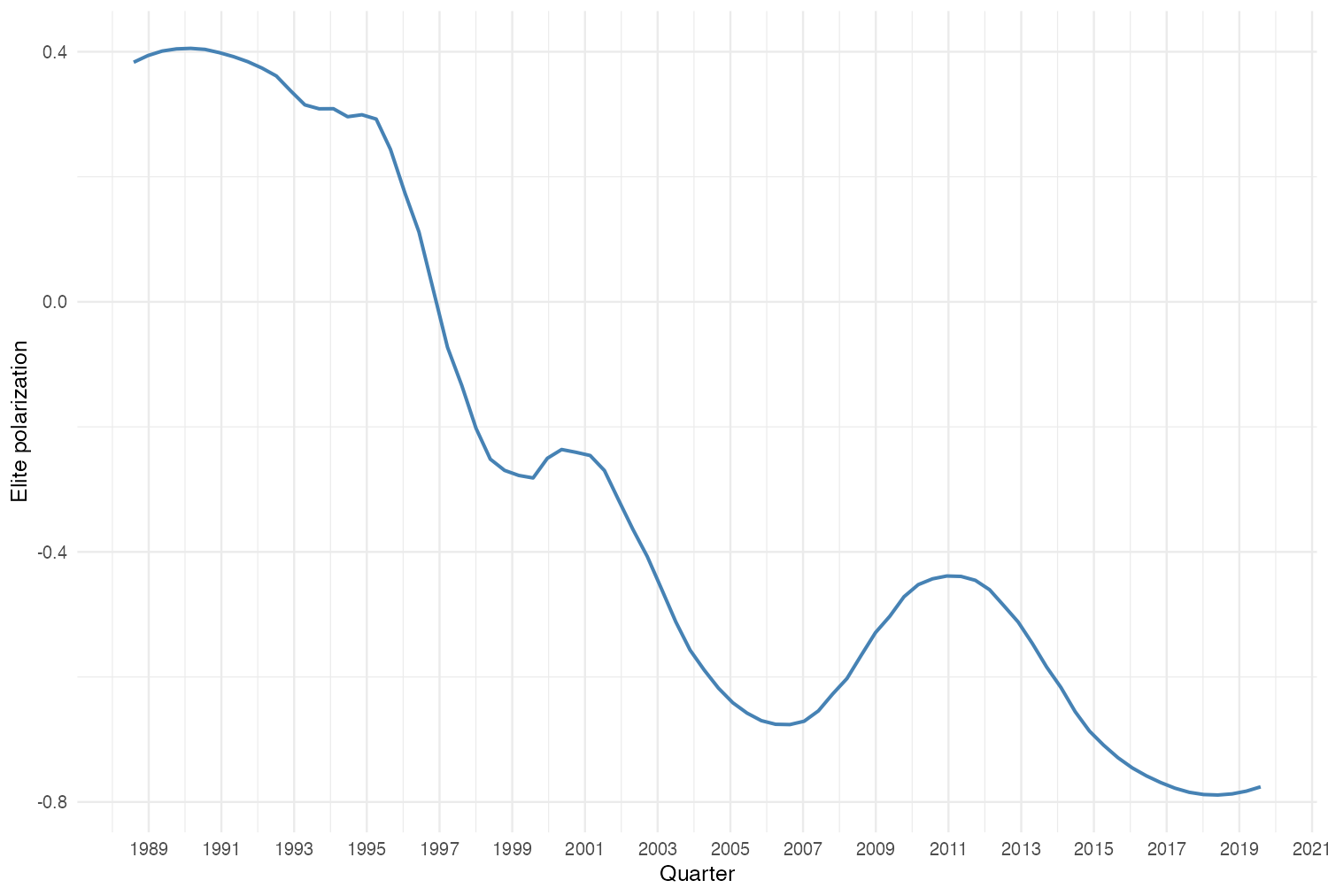}
\caption{UK elite polarization, 1987--2020\label{fig:overall_polarization_reiljan}}
\end{figure}

During the study period, the Conservatives governed until 1997, after which the New Labour held power until 2010, and then the Conservatives returned to office and remained there through the end of the sample window. Figure~\ref{fig:aps_uk_q} shows that Labour's out-party negativity fell sharply after New Labour entered government in 1997 and remained low until its 2010 defeat. The Conservatives also became less negative during the New Labour period. After 2010, however, the party pattern again resembles the broader expectation: the governing Conservatives record lower out-party negativity than the opposition Labour Party; losing power sharpened its critique. The aggregate elite polarization series in Figure~\ref{fig:overall_polarization_reiljan} therefore shows a U-shape with a post-1997 decline and post-2010 increase. The overall warming of the UK parliament rhetoric in the period under study is consistent with the general sentiment analysis \citep{ludovic2016}. 

The index responds to some electoral moments, especially the 1997 transition and the 2010 change of government, but not to every campaign or crisis. The 2001 campaign leaves little visible trace. The 2017 crisis was set against the backdrop of the ongoing Brexit debate; the loss of a clear majority severely hampered the government's ability to negotiate a definitive exit strategy with the European Union, ultimately leading to snap elections and a wave of debates. Yet the index does not show an increase in elite polarization during the Brexit period.
\begin{figure}[htbp]
\centering
\includegraphics[width=0.8\linewidth]{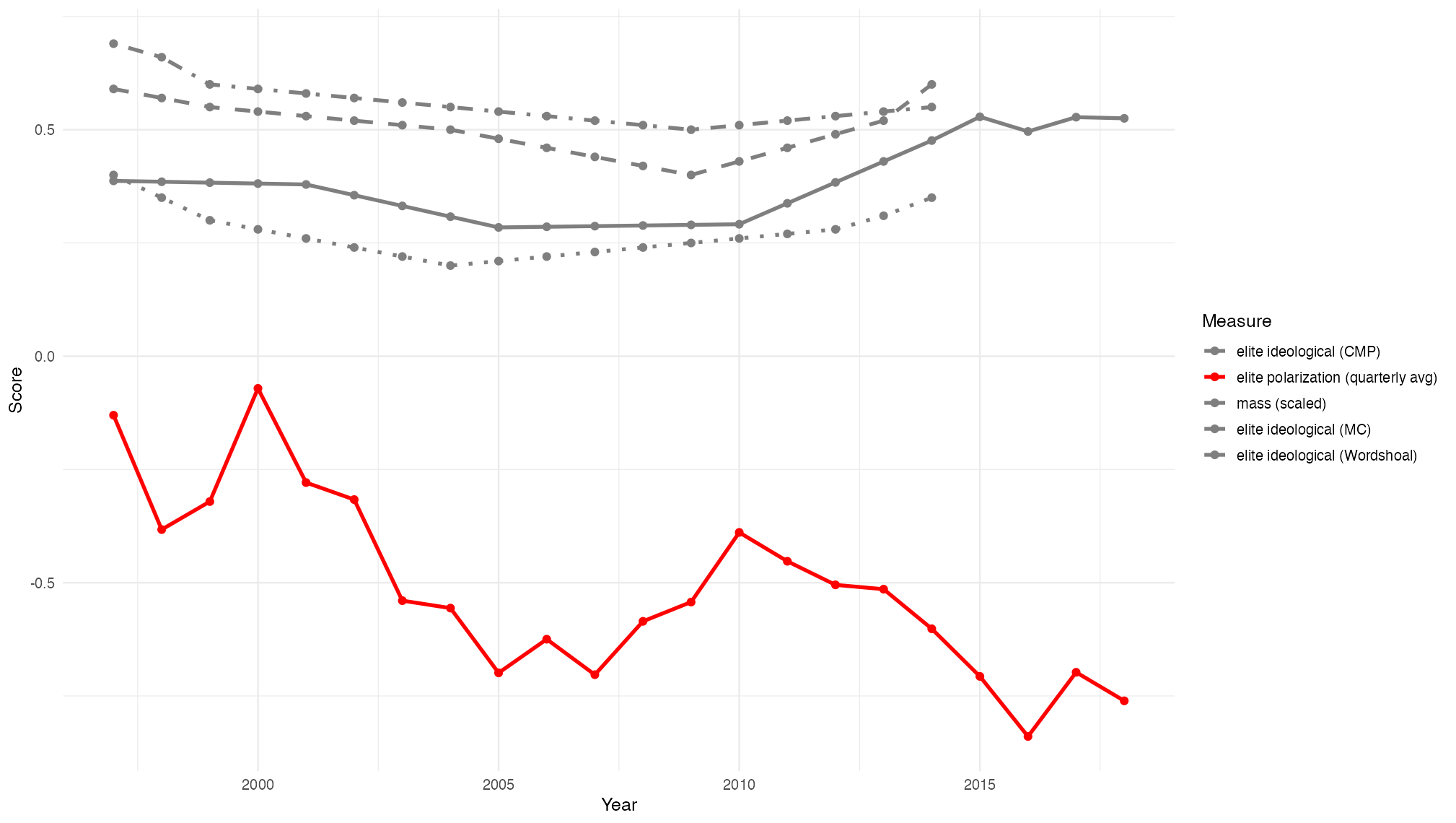}
\caption{UK elite polarization compared with mass and ideological measures, 1987--2020\label{fig:uk_three_polarizations}}
\end{figure}

Figure~\ref{fig:uk_three_polarizations} compares the elite polarization score with three existing NLP indices of UK parliamentary \emph{ideological} polarization aggregated by \citet{goet2019} (dictionary, supervised, and unsupervised methods), and with the UK mass-level \textit{affective} polarization series from \citet{reiljan2020}.\footnote{The Goet ideological indices end in 2014; The Reiljan mass-affective series ends in 2018;  the figure plots them only over their actual data range. The elite polarization series continues through the end of the sample window.} 

Against the ideological indices, the level correlation from 1989 to 2014 is $r = 0.92$, but this almost entirely reflects a shared downward trend across the New Labour decade: detrended (linear year residuals), the correlation falls to $r = 0.68$. Splitting the window separates two regimes. From 1997 to 2010, the two families of indices move together as British parties converge on the New Labour centre ($r = 0.74$, $n = 14$). After 2010, they decouple sharply: the ideological indices begin to rise again through the early Cameron years while elite polarization score continues to fall ($r = -0.97$ over $2010-2014, n = 5$). The indicator behaves in a similar way against the Reiljan mass-affective series. UK mass affective polarization slightly decreases between 2001 and 2010, together with my measure of elite polarization ($r = 0.75 $ over $2001, 2005, 2010, n = 3$). Similarly, after 2010, mass and elite affective polarization in the UK are moving in opposite directions ($r=-1$  over $2010, 2015, 2017$, $n = 3$): ideologies diverge, partisans grow more critical of the out-party, while their MPs do not.

Elite polarization proves to be a distinct construct. It co-moves with elite ideological polarization when ideological convergence is itself a story of dampened inter-party rivalry, the New Labour decade, but the two decouple and move in opposite directions, when ideological distance widens without correspondingly hostile rhetoric. It is also distinct from mass affective polarization: the parliamentary chamber and the electorate are not on the same trajectory in the United Kingdom over this period.

\subsection{Elite Polarization in Italy in the years 2013-2022}
Italy from 2013 to 2022 constitutes a case of medium to low level of affective mass polarization \citep[p.~9]{orhan2022}, yet quite volatile, highly fragmented proportional party system.  The second half of that decade is characterized by a chronic search for stable coalitions \citep[p.~63]{chiaramonte2017}: Renzi's center-left reforms and referendum defeat in 2016, the M5S–Lega "yellow-green" experiment in 2018–19, the M5S–PD "yellow-red" cabinet in 2019–21, and the technocratic Draghi unity government in 2021–22.  Each turnover forced former rivals into an alliance. In the Italian case, parties' out-party negativity fluctuated with them entering or leaving the government.

From 2013 to early 2018, the Democratic Party (PD) governed in a grand coalition with the center-right and then under successive center-left cabinets; the Five Star Movement (M5S) remained outside government and expressed the highest out-party negativity among the three parties plotted here. PD was generally less negative while it governed. During this period, the aggregate Italian score remains positive and comparatively stable, with a high point around 2015--2017.

\begin{figure}[htbp]
\centering
\includegraphics[width=0.8\linewidth]{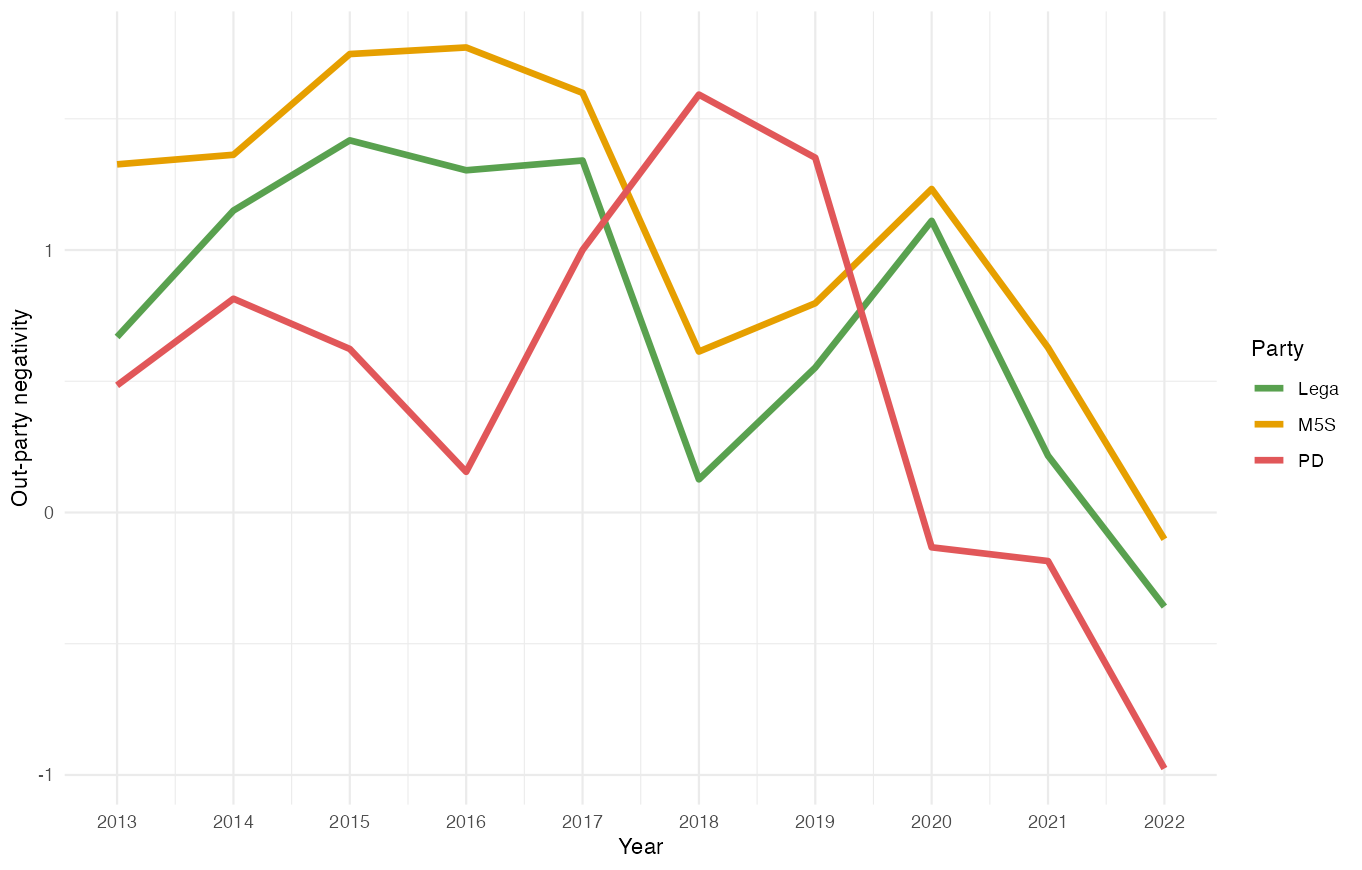}
\caption{Out-party negativity in Italy, by party and year, 2013--2022\label{fig:aps_it_coal}}
\end{figure}

\begin{figure}[htbp]
\centering
\includegraphics[width=0.8\linewidth]{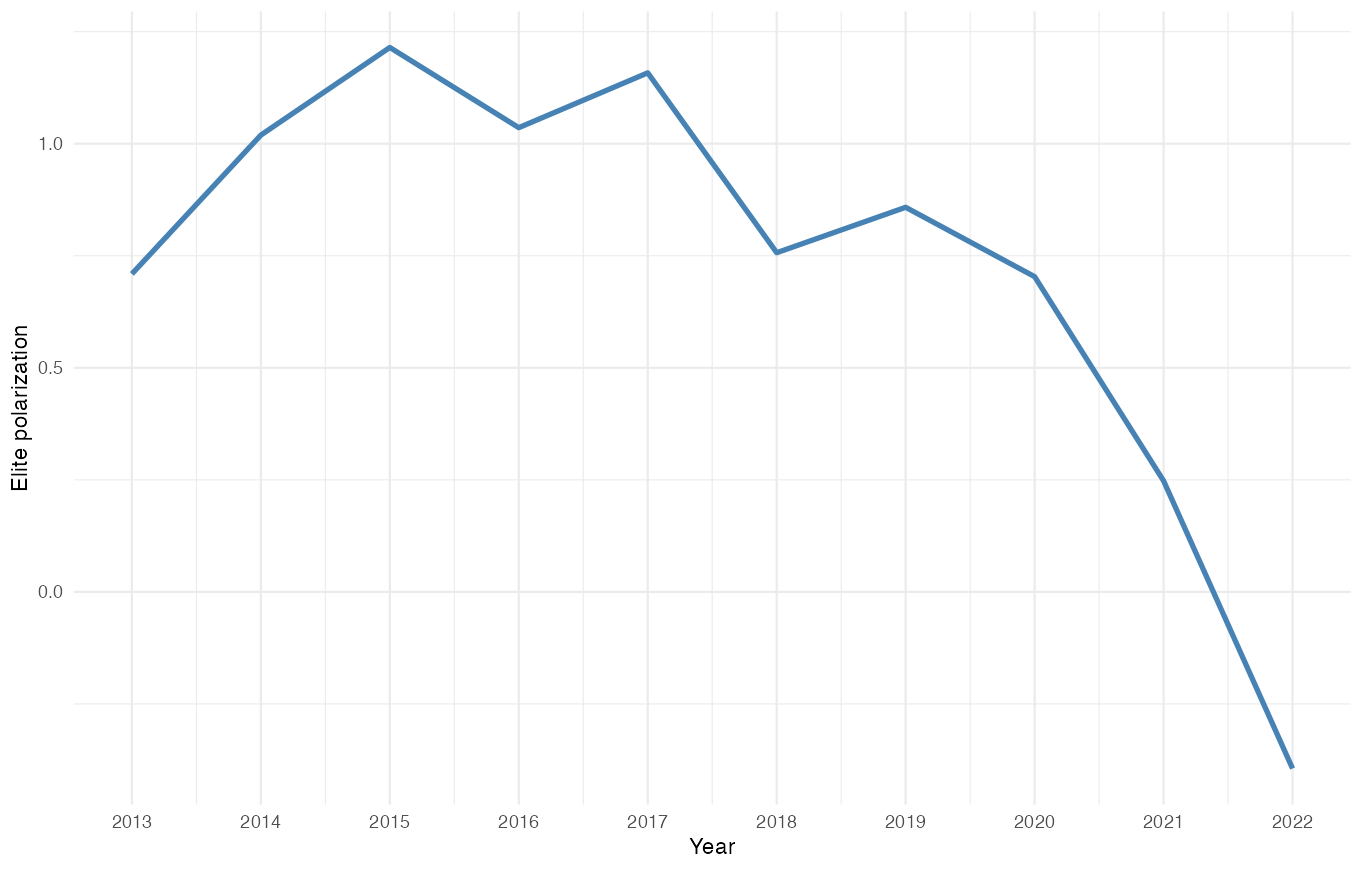}
\caption{Elite polarization in Italy, 2013--2022\label{fig:aps_it_overall}}
\end{figure}

The 2018 election is visible in the series, but not as a simple polarization spike. At the aggregate level, Figure~\ref{fig:aps_it_overall} shows a decline in elite polarization after 2017. At the party level, Figure~\ref{fig:aps_it_coal} shows why: M5S and Lega lowered their out-party negativity as they entered the ``yellow-green'' government, while PD, excluded in this 2018--2019 period, became the most negative of the plotted parties, consistent with the existing literature \citep[p.~7-9]{giannetti2020}. When the coalition collapsed in August 2019, Lega moved to the opposition and its critique of out-parties rose in the following year, while M5S formed a new ``yellow-red'' government with PD. Because the graph is yearly, 2019 still mixes the transition itself with the first months of the new cabinet; the clearer decline in PD and M5S out-party negativity appears after 2020. In February 2021, almost all major parties---including PD, M5S, and Lega---entered the Draghi national-unity cabinet, producing the lowest elite polarization in the series and concentrating elite hostility on the sole opposition party, Brothers of Italy (not included in the graph) \citep[p.~173]{russo2022}.

Over the decade under study, Italian elite polarization generally rises when parties move into opposition and falls when they share cabinet responsibility. This speaks for the validity of the index, while the 2018 case also shows how elections can reduce measured parliamentary negativity when former opponents become governing partners.

\section{Conclusion}

This study has introduced the Elite Polarization Score---the cross-national, time-series measure of out-party hostility drawn directly from parliamentary speech. The methodological core of the approach is a shift in the unit of analysis: rather than assigning sentiment to whole speeches, the LLM-powered pipeline identifies every instance in which one politician evaluates their counterpart, attributing valence at the level of the individual reference. This disaggregation resolves a long-standing bottleneck in political science NLP, enabling reliable sentiment attribution to specific political dyads at scale. Because LLMs transfer across languages without task-specific retraining, the same pipeline operates on Hungarian, Italian, and English corpora without additional adaptation — making a genuinely multilingual, longitudinal panel tractable. Conceptually, the resulting measure occupies the previously empty elite-affect corner of the polarization typology: it is distinct from ideological distance between parties, and distinct from the mass-level partisan hostility captured by survey thermometers, targeting instead the observable rhetorical behavior of legislative elites toward one another.

Leveraging state of the art LLMs boosts accuracy and sensitivity well beyond earlier rule-based or supervised techniques. The validation protocol shows that the mean absolute error in sentiment has narrowed to 0.85--1.48 points on a 10-point scale, while AI sensitivity meets or exceeds that of a human coder, and false-discovery rates are negligible. These results signal that most off-the-shelf LLMs are capable of fine-grained political-text analysis, which once required fine-tuning or even training.

The case studies have shown the high face validity of the index, while also clarifying what kind of political change it captures. Government participation is often associated with lower out-party negativity: New Labour becomes much less negative after entering office in 1997, the post-2010 Conservatives also record low levels while governing, and Italian parties tend to become less negative when they enter cabinet coalitions and more negative when they leave them. Hungary is more complex: the aggregate series remains moderate during the first phase of Fidesz's executive aggrandizement and peaks later, after the party system and opposition structure had already changed. At the same time, the index is not a generic crisis detector. It does not show a major elite-polarization spike during the Brexit period or the 2001 UK campaign, and it reacts only weakly to the Hungarian 2006 crisis. In Italy, by contrast, the 2018 election is visible, but as a reduction in aggregate elite polarization caused by former opponents entering government together rather than as an immediate hostile spike. These limitations are consistent with the fact that parliamentary speech is choreographed and institutionally constrained.

The immediate next step is to extend the validated pipeline to the remaining EU member states, producing a twenty-year, EU-wide panel of elite polarization. The corpora for most member states are available through the ParlaMint project \citep{erjavec2023}, which now covers twenty-nine national parliaments, and the per-speech annotation cost is low enough that a full EU run is computationally tractable. The most promising operational route, as flagged earlier in the NLP section, is to execute this extension in \textit{agentic} mode: an orchestrator LLM autonomously dispatches corpus retrieval, actor-subject detection, sentiment scoring, fuzzy matching, and quality checks across all twenty-nine parliaments in parallel, with human intervention reserved for adjudication of flagged cases rather than routine coding \citep{wangsurvey2024}. Under such a regime the binding constraint shifts from researcher labor to compute budget, and what would otherwise be a year-long annotation effort can in principle be compressed into weeks.

Beyond panel construction, the EPS opens several routes for future research. First, it would allow scholars to study elite polarization as a dependent variable: whether it grows gradually in response to structural features of party systems, or whether it reacts sharply to elections, crises, scandals, coalition breakdowns, and other political shocks. The index also makes it possible to compare pre- and post-election periods, to test whether government and opposition parties polarize differently, and to examine whether the degree of out-party hostility varies by topic of debate. Second, the measure enables discriminant validity tests against neighbouring forms of polarization. Future work can compare the Elite Polarization Score to survey-based mass affective polarization indices \citep{reiljan2020, orhan2022}, manifesto-based ideological distance measures, and existing parliamentary ideology scores, asking when elite affective hostility moves together with, or separately from, mass affect and ideological distance. Third, the EPS can be used as an independent variable in models of elite-driven political change, including populist escalation, democratic norm erosion, institutional hardball, and democratic backsliding. In this sense, the measure makes it possible to ask not only what produces elite polarization, but also what elite polarization itself produces.

Deriving from the algorithms offered in this paper, political science scholars can study other aspects of relations in elite speeches: for instance,  in competitive authoritarian regimes, highlighting business clientele, support for the events of recent executive aggrandizement, or rhetorical support of the incumbent; in democracies, lobbying, or within-party competition in dominant party regimes. By demonstrating that LLM-assisted text analysis can deliver theory-relevant, cross-national measures at scale, the project opens the door to a new generation of comparative-politics datasets. In addition, this project demonstrates many other ways of using pre-trained LLMs: classifying political actors, fuzzy matching the long tables with the mentions of politicians, and machine translating the speeches for coders.

    \bibliography{references}

    \appendix
    \setcounter{table}{0}
    \setcounter{figure}{0}
    \renewcommand{\thetable}{A\arabic{table}}
    \renewcommand{\thefigure}{A\arabic{figure}}
    \renewcommand{\theHtable}{appendix.A\arabic{table}}
    \renewcommand{\theHfigure}{appendix.A\arabic{figure}}

    \section{Data Sources and Coverage}

    \begin{table}[htbp]
    \centering
    \caption{Corpus Coverage in the Current Three-Country Demonstration}
    \begin{tabularx}{\linewidth}{llX}
    \toprule
    Country & Period & Main source and current repaired table \\
    \midrule
    Hungary & 2002--2023 &
    ParlLawSpeech and ParlaMint-HU \\
    United Kingdom & 1988--2020 &
    House of Commons V2 / ParlSpeech-derived corpus \\
    Italy & 2013--2022 &
    ParlaMint-IT \\
    \bottomrule
    \end{tabularx}
    \end{table}

   \begin{table}[htbp]
\centering
\caption{Reference extraction coverage in repaired mention files\label{tab:reference_coverage}}
\small
 \begin{tabularx}{\linewidth}{lllllX}
\toprule
Country & Speeches & Any refs & Party-affiliated refs & Party-affiliated \\
\midrule
Hungary & 12,400 & 55,962 & 16,499 & 29.5\% \\
United Kingdom & 458,752 & 2,453,301 & 1,593,312 & 64.9\% \\
Italy & 32,459 & 167,316 & 31,116 & 18.6\% \\
\bottomrule
\end{tabularx}
\begin{flushleft}\footnotesize Notes: Speeches are distinct speech identifiers with at least one extracted reference in the repaired mention table. Party-affiliated references are extracted references whose referent affiliation resolves to a party label, excluding generic government, institutional, foreign-institution, independent, other, unknown, and undefinable labels.\end{flushleft}
\end{table}

    \section{Extreme Negative Evaluations and Pernicious Rhetoric}

    Table~\ref{tab:incivility_examples} reports the share of the most negative and moderately negative references that are manually coded as pernicious or incivility-like. The example rows are selected deterministically from the coded source sheets.

\begin{table}[htbp]
\centering
\caption{Extreme negative evaluations and pernicious rhetoric\label{tab:incivility_examples}}
\scriptsize
\begin{flushleft}\footnotesize \textit{Summary:} Each sentiment bin contains 100 coded references. Pernicious rhetoric appears in 27.8\% of the most negative sample ($-5$) and 3.0\% of the moderately negative comparison sample ($-2$).\end{flushleft}
\vspace{0.25em}
\begin{tabularx}{\linewidth}{@{}p{0.10\linewidth}p{0.17\linewidth}X@{}}
\toprule
Sample & Reference ID & Rationale \\
\midrule
-5 & 2017-07-13261 & Daesh is described as a callous enemy, responsible for heinous acts and ongoing threats to safety. The speaker's consistent framing of Daesh as a brutal organization that has caused extensive suffering and challenges positions them negatively. \\
-5 & 1993-11-01127 & The speaker condemns the actions of the Ulster Freedom Fighters in the context of violence and terrorism. The speaker expresses strong disapproval of the Ulster Freedom Fighters, labeling their actions as barbaric and despicable. \\
-5 & 2017-07-05410 & The speaker discusses breaches of international law and human rights abuses in the occupied Palestinian territories, reflecting a critical stance toward the situation there. "The Palestinian Territories represent a political region with significant international relations implications." \\
-2 & 1999-07-13294 & The IRA is mentioned when discussing concerns about violence and its impact on peace efforts, contrasting its actions with the desire for conflict resolution. The speaker conveys a negative sentiment towards the IRA, highlighting its role in intimidation and terrorism. \\
-2 & 1997-01-27155 & The term is specifically mentioned in the context of discussing their mental health and legal treatment. The speaker expresses a negative sentiment regarding paedophiles, suggesting they are mentally sick with no effective treatment. \\
-2 & 2001-11-0718 & Mentioned in context of terrorist activity attributed to them, highlighting their role in violence. The speaker expresses a negative sentiment as this group is associated with terrorism. \\
\bottomrule
\end{tabularx}
\begin{flushleft}\footnotesize Notes: Example rows are the first three coded pernicious cases in each sample by source-file order. Pernicious rhetoric is coded as references that delegitimize, existentially threaten, or otherwise cast the target as outside acceptable democratic conflict.\end{flushleft}
\end{table}

    \section{Validation Protocol}

    \subsection{Verification protocol for the human assistant}
    Here is what human coders are instructed to do.
    ``Take a random sample of non-consecutive texts from the source database. Then strictly follow the text of the Prompt, which I also give to the artificial intelligence.''
    Then follows the prompt text, and the examples of ChatGPT 4's built-in sentiments to anchor their scoring.
    After the coding is finished, the following protocol is used for checking the sensitivity of finding at the script stage.
    \begin{enumerate}
        \item Run the AI matching model on the same batch of 10 texts.
        \item In the file that matches the findings of AI and those of human coders, correct duplicates and errors in the matching.
        \item Measure performance in the sentiment using the resulting file.
    \end{enumerate}

    \subsection{Examples of ChatGPT 4's built-in sentiments}
    $-5$: N is a complete disgrace; his actions are unforgivable and heinous.\\
    $-4$: N's policy is catastrophic and shows her incompetence and lack of empathy.\\
    $-3$: N's approach is flawed and disappointing; he must reconsider his stance.\\
    $-2$: N's recent decision was short-sighted and not well-thought-out.\\
    $-1$: N could have done better; the results were somewhat underwhelming.\\
    $0$: N presented the quarterly report as per the schedule.\\
    $+1$: N made a decent effort, and there are some signs of improvement.\\
    $+2$: N's initiative is somewhat effective; it's a step in the right direction.\\
    $+3$: N's leadership has been strong and beneficial to our team.\\
    $+4$: N's contribution has been outstanding and transformative for our project.\\
    $+5$: N is an extraordinary visionary; their work has revolutionized our understanding and approach.

    \section{Robustness Checks}
    
\begin{figure}[htbp]
\centering
\includegraphics[width=\linewidth]{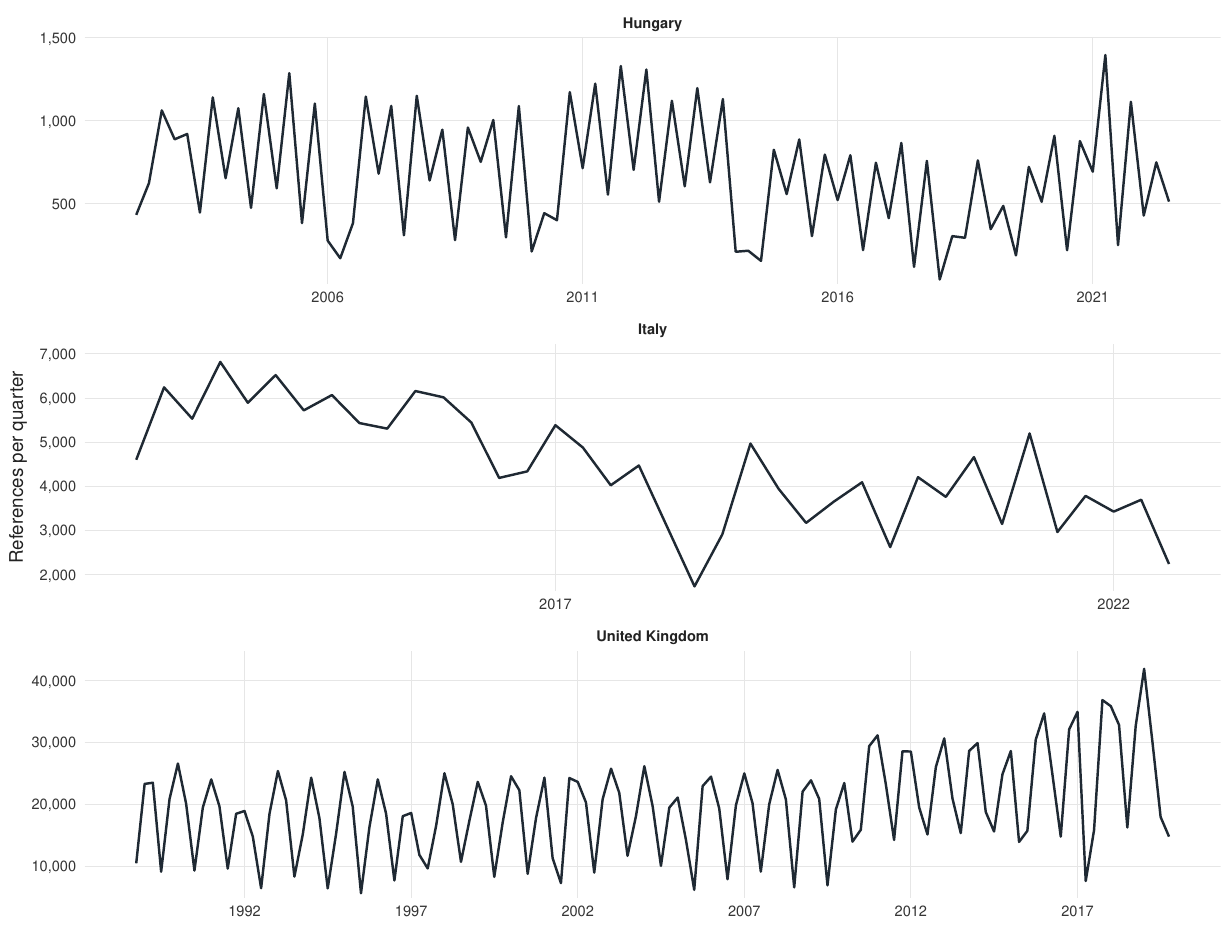}
\caption{Reference volume over time by country. The diagnostic checks whether
changes in elite polarization coincide with sharp changes in the number of
usable political-actor references. Country panels use separate vertical scales
for legibility.}
\end{figure}

\begin{figure}[htbp]
\centering
\includegraphics[width=\linewidth]{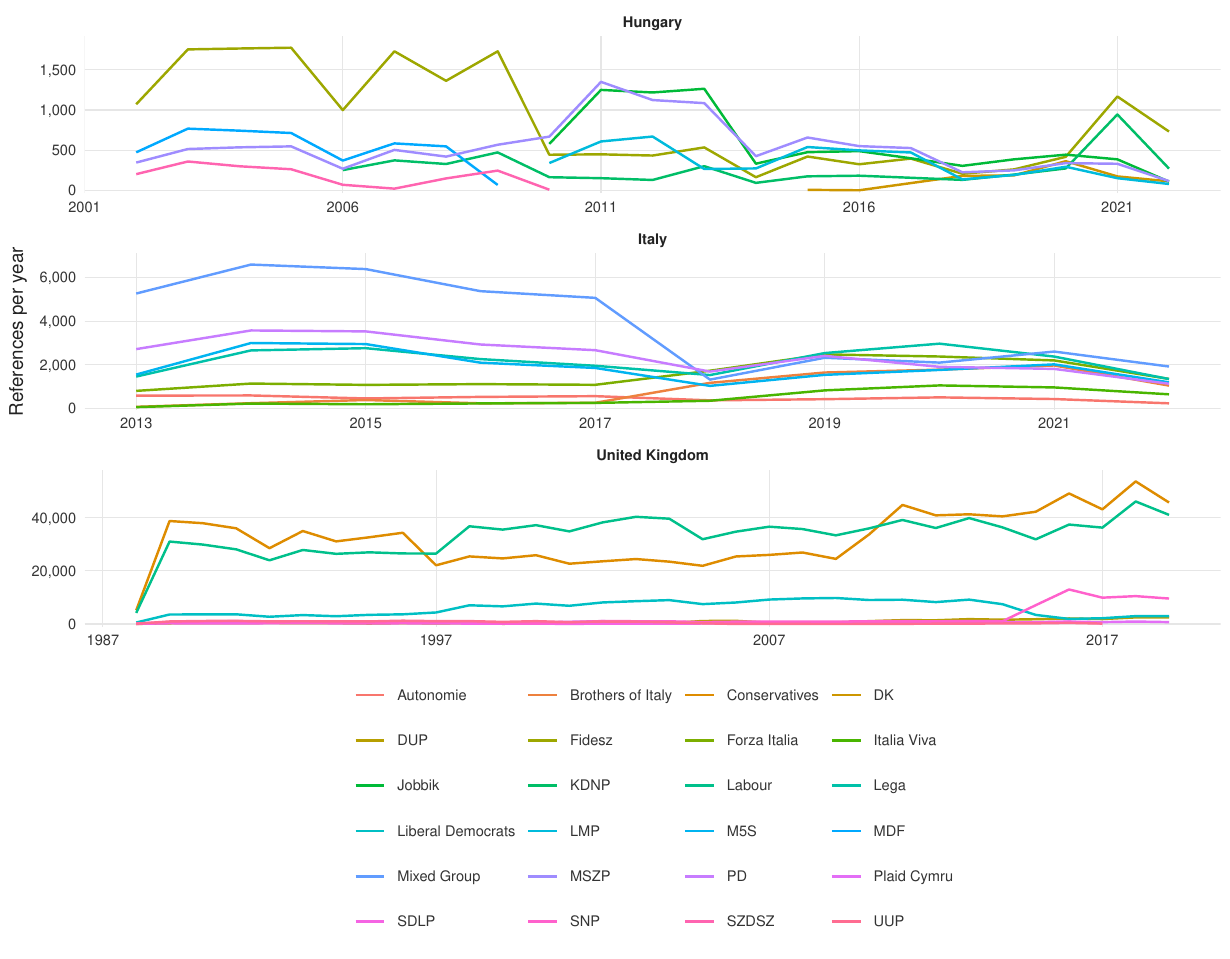}
\caption{Reference volume over time by major speaker party. The diagnostic
flags periods in which party-level out-party negativity may be supported by
thin or highly uneven mention volume. Country panels use separate vertical
scales for legibility.}
\end{figure}

\begin{figure}[htbp]
\centering
\includegraphics[width=\linewidth]{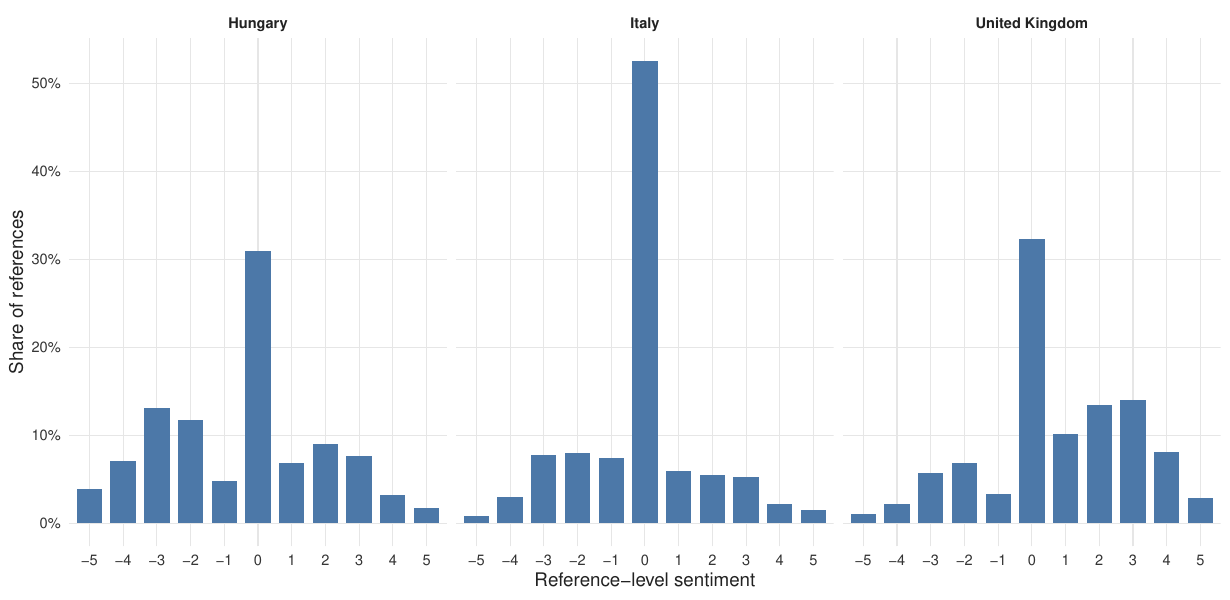}
\caption{Raw reference-level sentiment distributions by country. The diagnostic
checks whether cross-country differences are associated with different use of
the sentiment scale.}
\end{figure}

\begin{figure}[htbp]
\centering
\includegraphics[width=\linewidth]{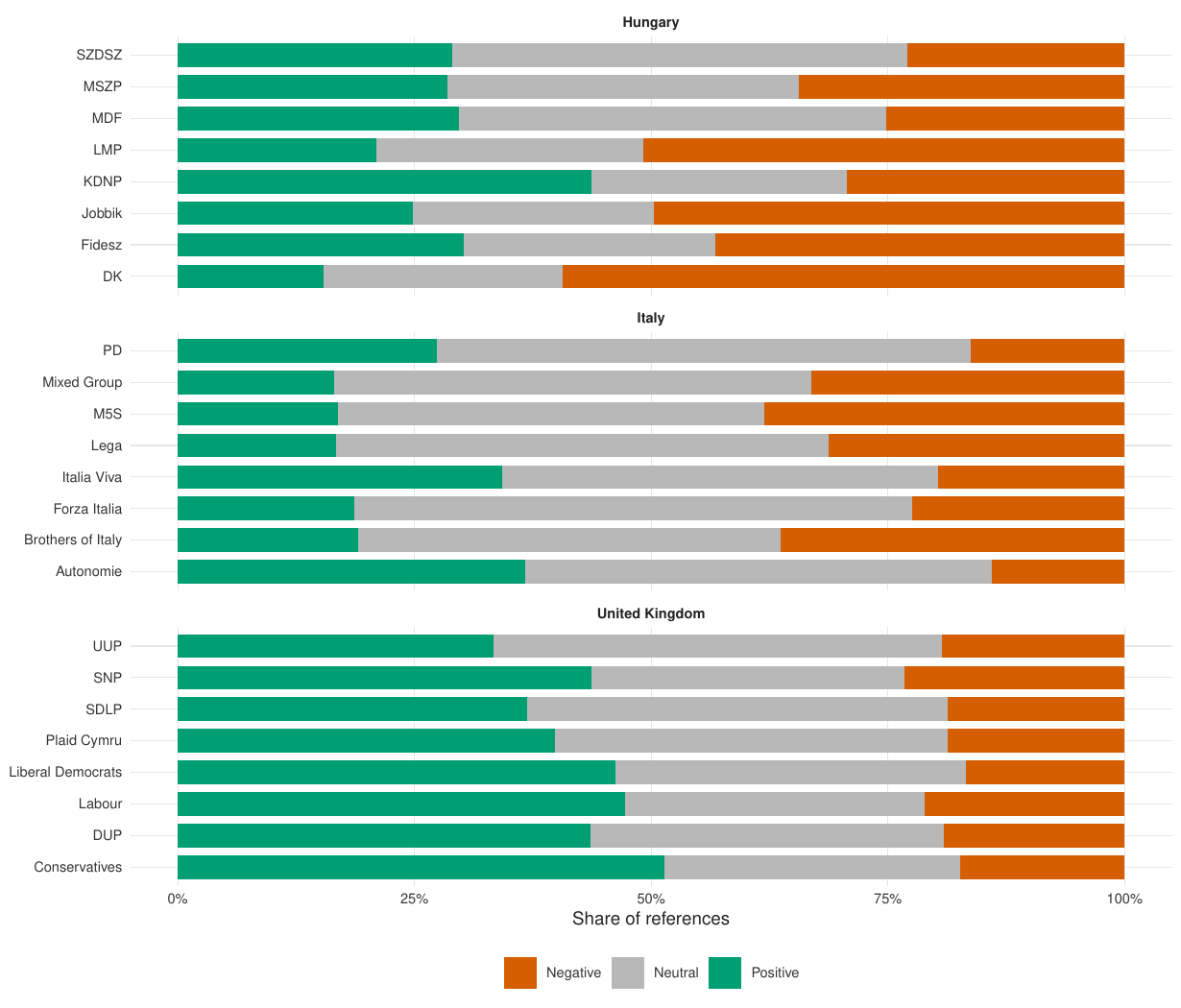}
\caption{Negative, neutral, and positive reference shares by major speaker
party. The diagnostic shows which parties contribute most to the negative and
positive tails of the sentiment distribution.}
\end{figure}

\begin{figure}[htbp]
\centering
\includegraphics[width=\linewidth]{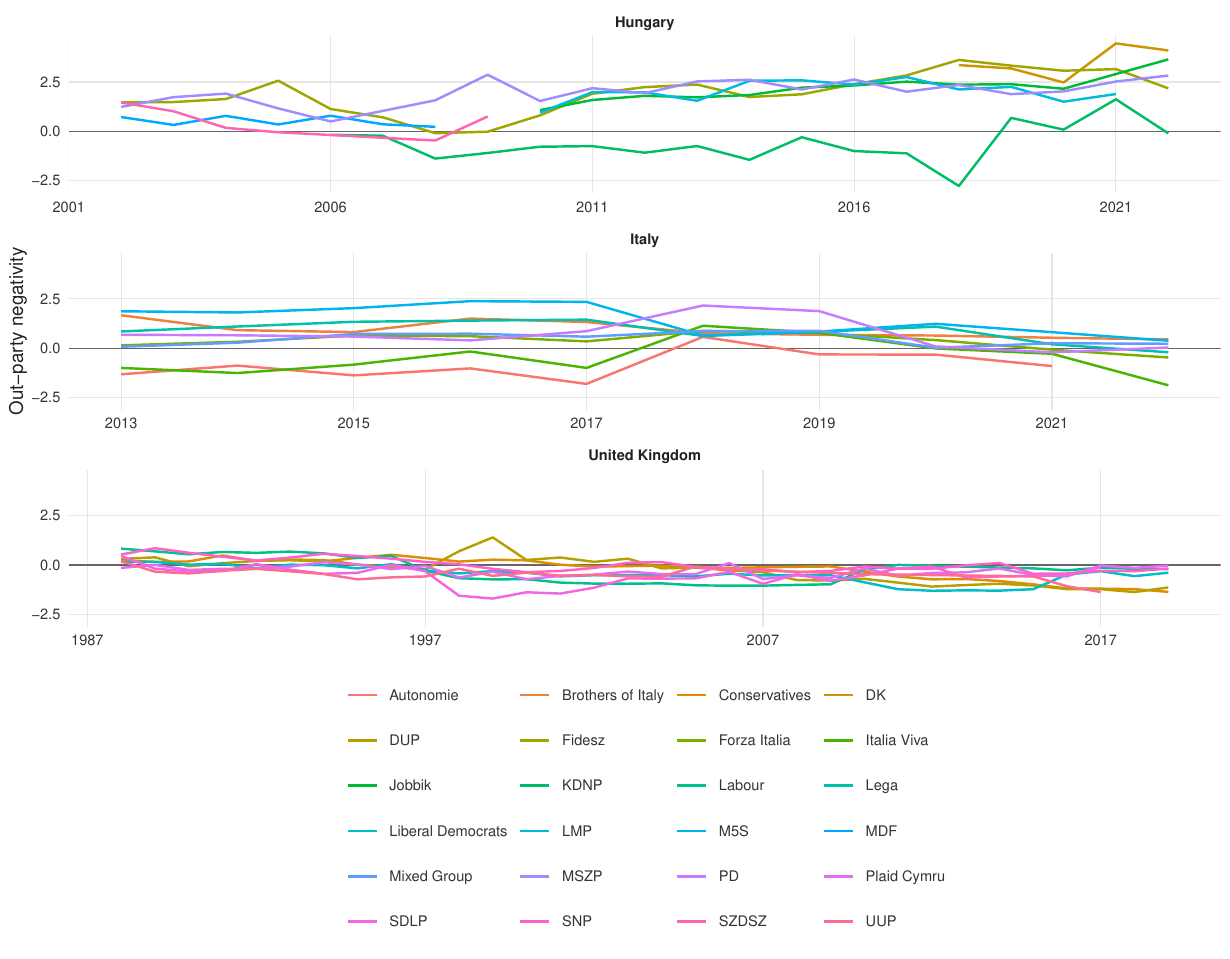}
\caption{All-party party-level out-party negativity. This diagnostic extends
the main party plots beyond the parties shown in the article. It uses
domestic-party referents among the most frequently observed speaker parties and
excludes party-year cells with fewer than ten domestic out-party references.}
\end{figure}

\begin{figure}[htbp]
\centering
\includegraphics[width=\linewidth]{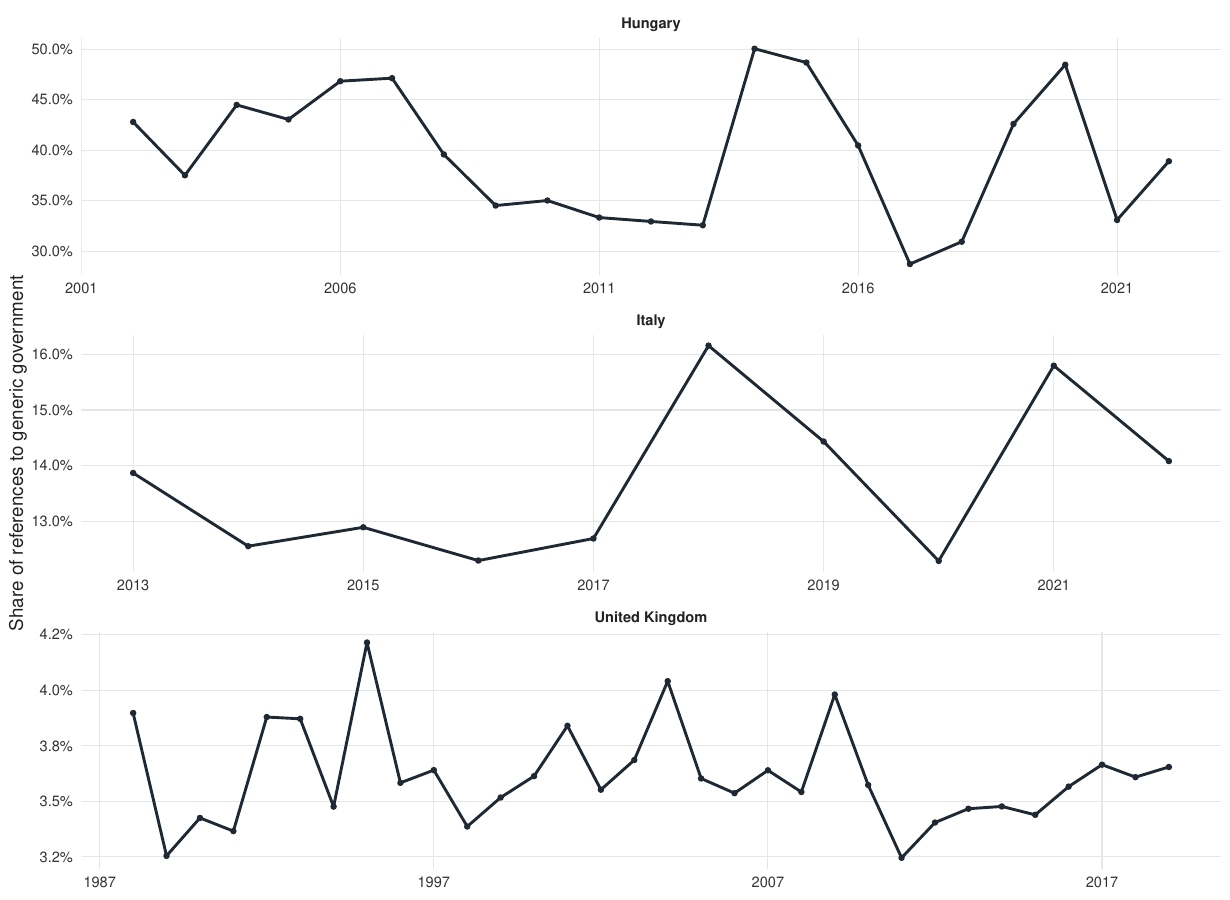}
\caption{Share of references made to generic government actors. The diagnostic
shows when government-as-party recoding matters most for party-level and
country-level results. Country panels use separate vertical scales for
readability.}
\end{figure}

\begin{figure}[htbp]
\centering
\includegraphics[width=\linewidth]{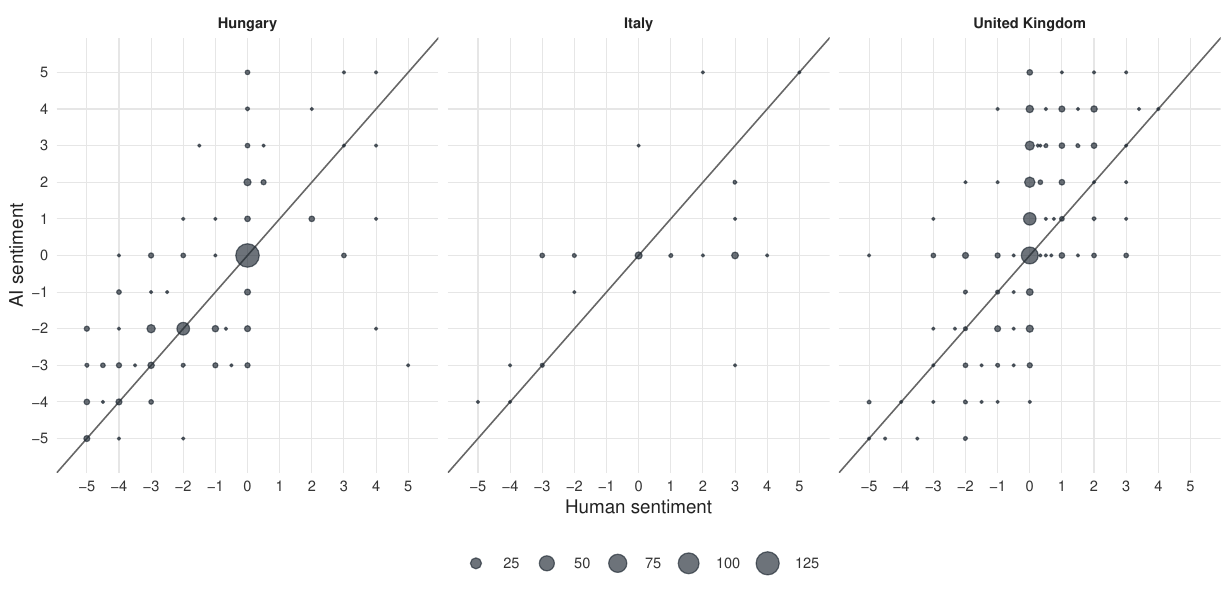}
\caption{Human-AI sentiment agreement for matched validation references. The
diagonal line marks exact agreement on the $[-5,+5]$ scale.}
\end{figure}

\begin{figure}[htbp]
\centering
\includegraphics[width=\linewidth]{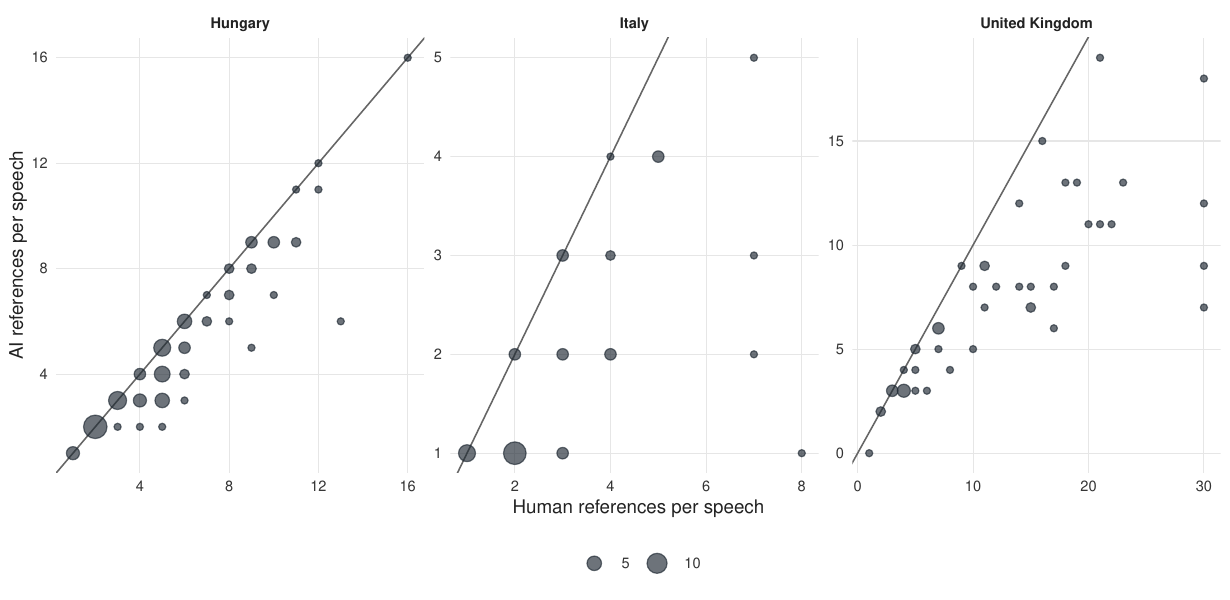}
\caption{Human-AI reference count agreement by validation speech. The diagonal
line marks equal numbers of references identified by the human coder and the AI
pipeline. Speeches with more than 30 references by either coder are omitted
from the plotted diagnostic for readability.}
\end{figure}

\begin{figure}[htbp]
\centering
\includegraphics[width=\linewidth]{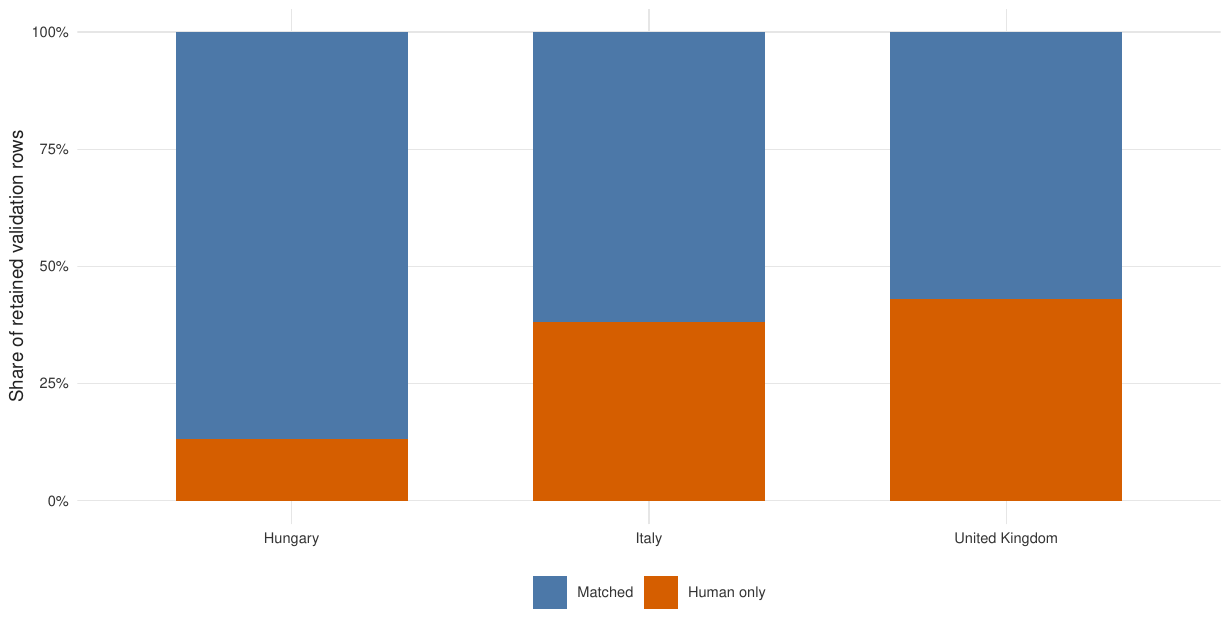}
\caption{Matched and human-only rows in the validation supergold files. The
diagnostic focuses on references that appear in the human-coded validation
material.}
\end{figure}

\FloatBarrier

\section{In-Group Reference Variant}

As discussed in the article, the main Elite Polarization Score deliberately uses out-party evaluation only. In-party references are still theoretically important, as they are used in most of the definitions of affective mass polarization. The graphs below show the option of an index that use 
\begin{figure}[htbp]
\centering
\includegraphics[width=\linewidth]{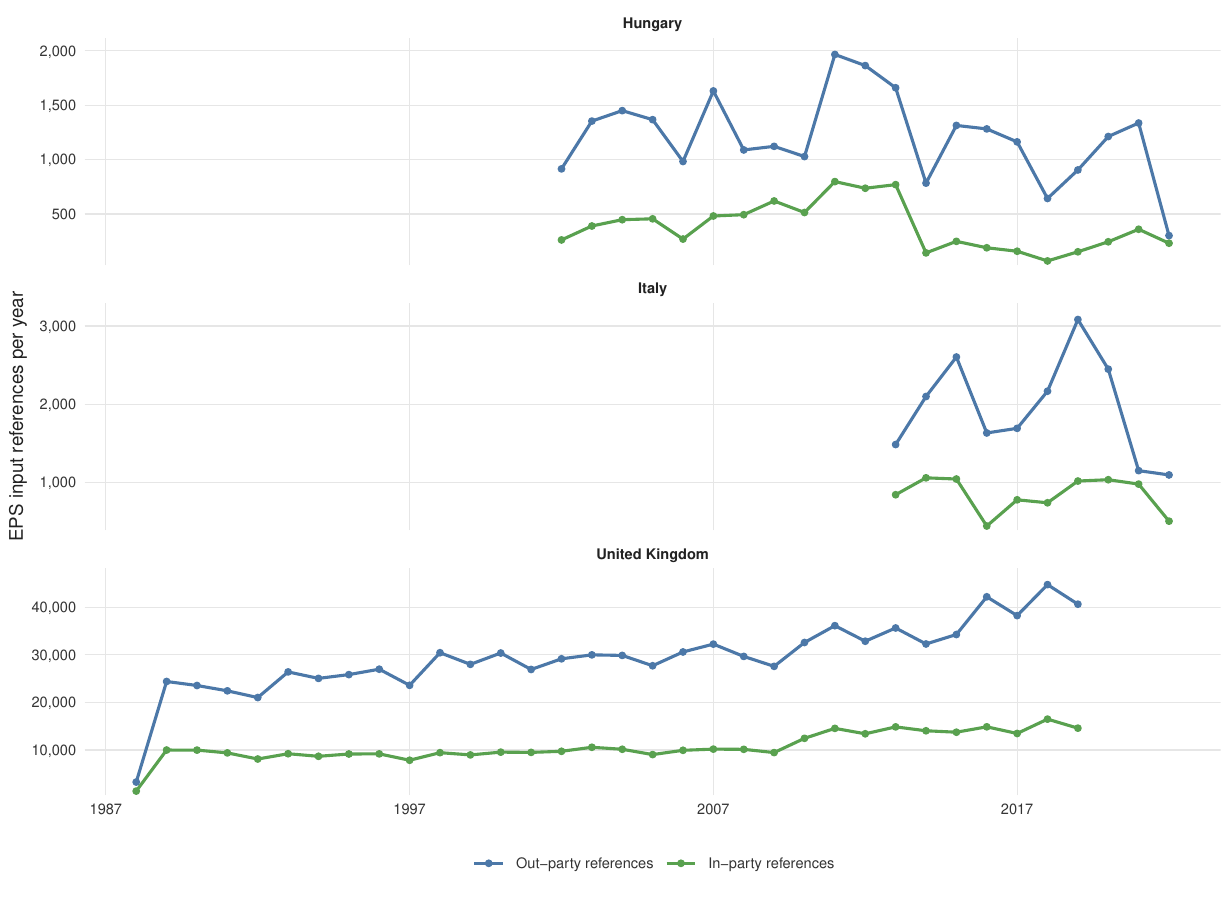}
\caption{In-party and out-party raw references per year in the EPS input data. Country panels use
separate vertical scales for readability.}
\end{figure}

\begin{figure}[htbp]
\centering
\includegraphics[width=\linewidth]{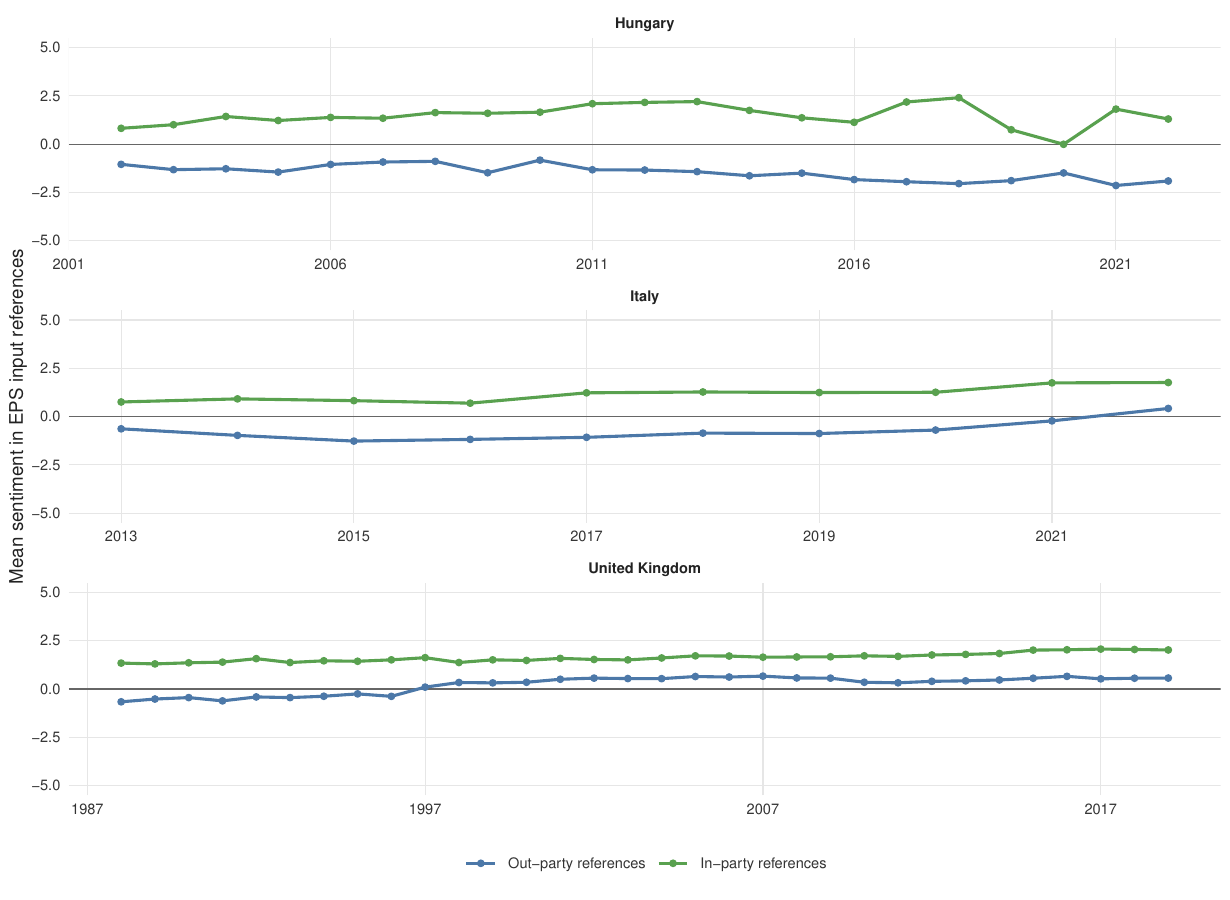}
\caption{Mean sentiment per ref-type and year in the EPS input data. The
series are near-stationary by construction. With tens of thousands of
references per country-year (about 70{,}000/year for the United Kingdom),
the standard error of the yearly mean on the $\pm5$ scale is below
$0.05$, so the average traces the production model's expected score per
ref-type rather than political change.}
\end{figure}

\begin{figure}[htbp]
\centering
\includegraphics[width=\linewidth]{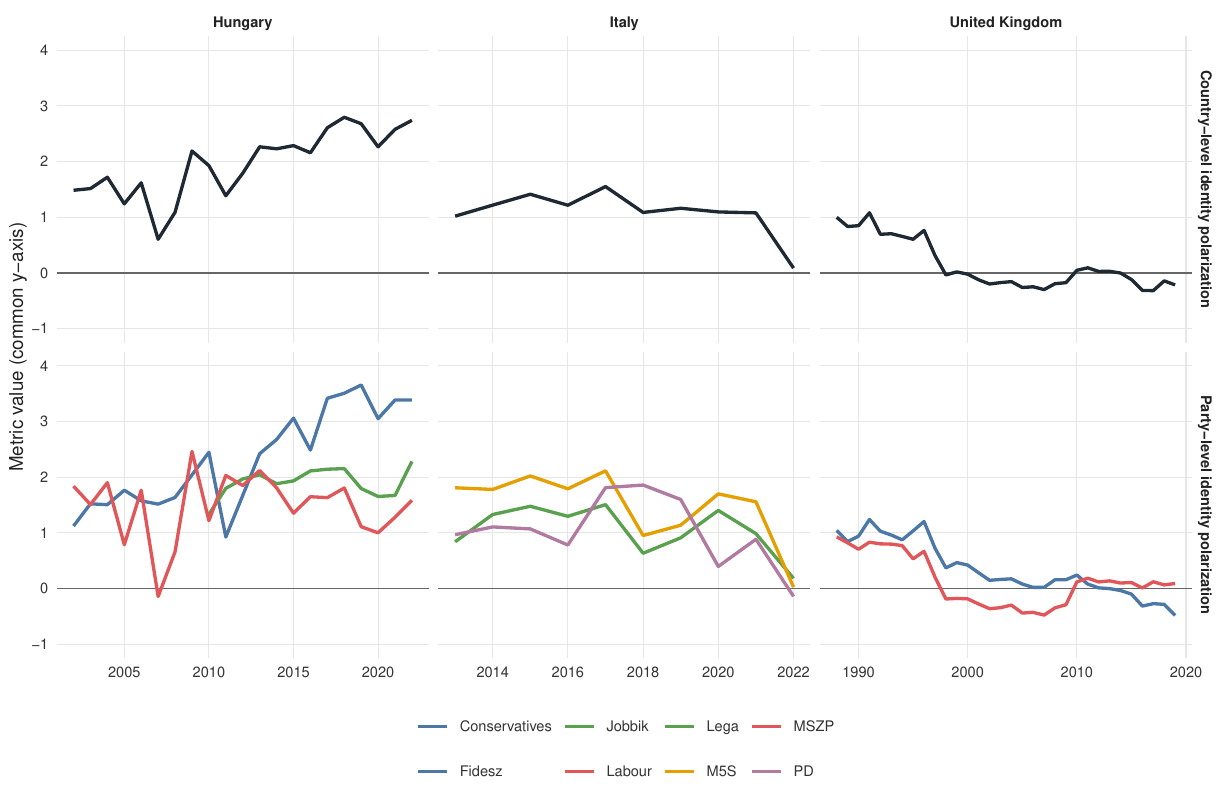}
\caption{Exploratory Elite Polarization Score option diagnostic with positive in-party references added to out-party negativity. The top row reports country-level
vote-share weighted values; the bottom row reports party-level values for the main parties. All panels use common vertical limits so that one unit occupies the same visual distance in every country panel.}
\end{figure}

\begin{figure}[htbp]
\centering
\includegraphics[width=\linewidth]{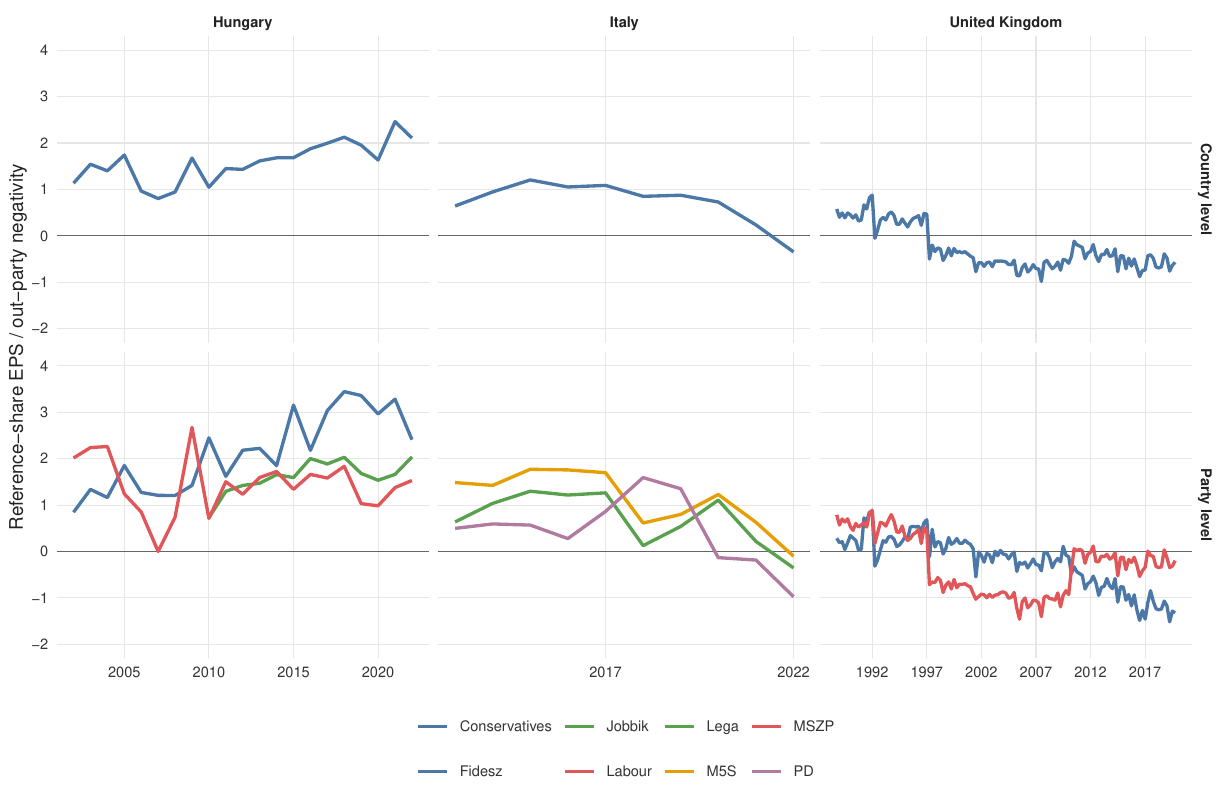}
\caption{Reference-share version of the Elite Polarization Score. The top row reports the country-level score, aggregating parties by their share of eligible
speaker-party references in the same period. The bottom row reports the corresponding party-level out-party negativity for the main parties. The common vertical scale keeps the units visually comparable across countries.}
\end{figure}

\begin{figure}[htbp]
\centering
\includegraphics[width=\linewidth]{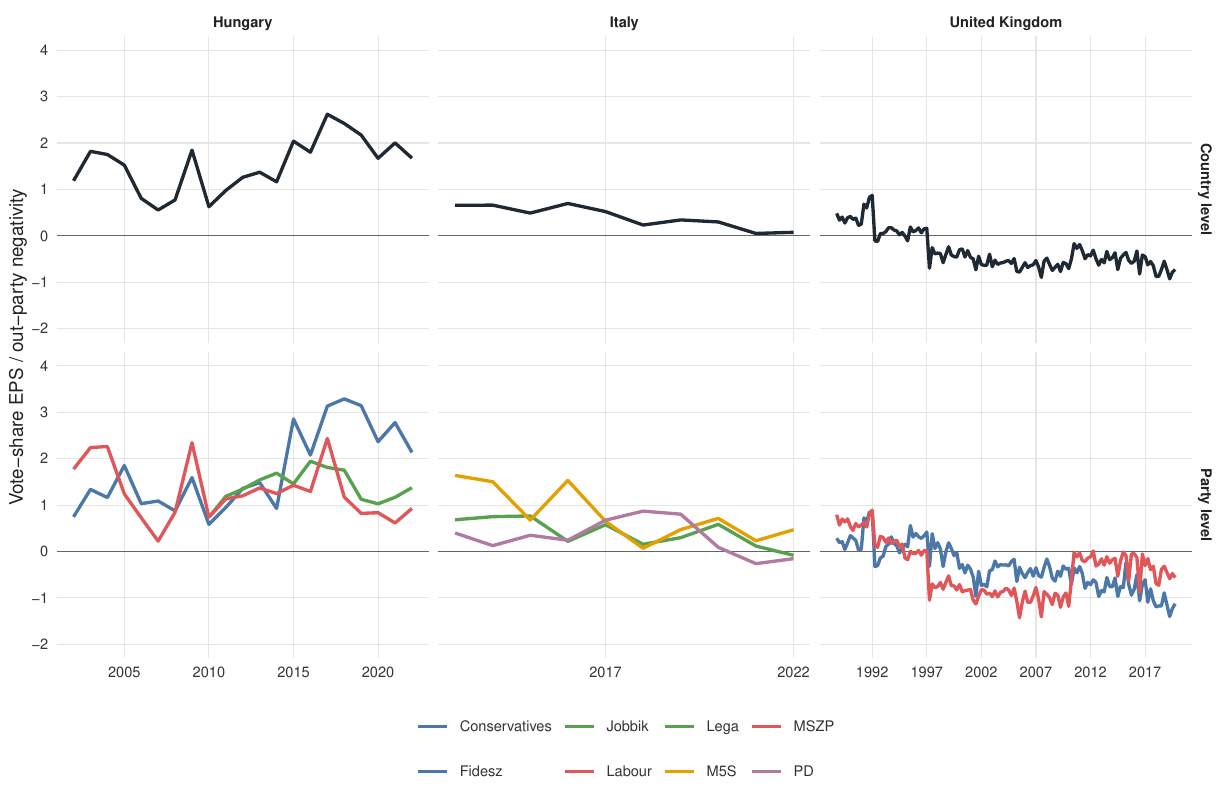}
\caption{Vote-share version of the Elite Polarization Score. The top row reports the country-level score, aggregating speaker parties by vote share. The bottom row reports party-level out-party negativity when out party targets are weighted by electoral vote share rather than by reference share. Periods with no eligible mention data are not plotted.}
\end{figure}

\section{EPS share-vs-sentiment decomposition}
\begin{figure}[htbp]
\centering
\includegraphics[width=\linewidth]{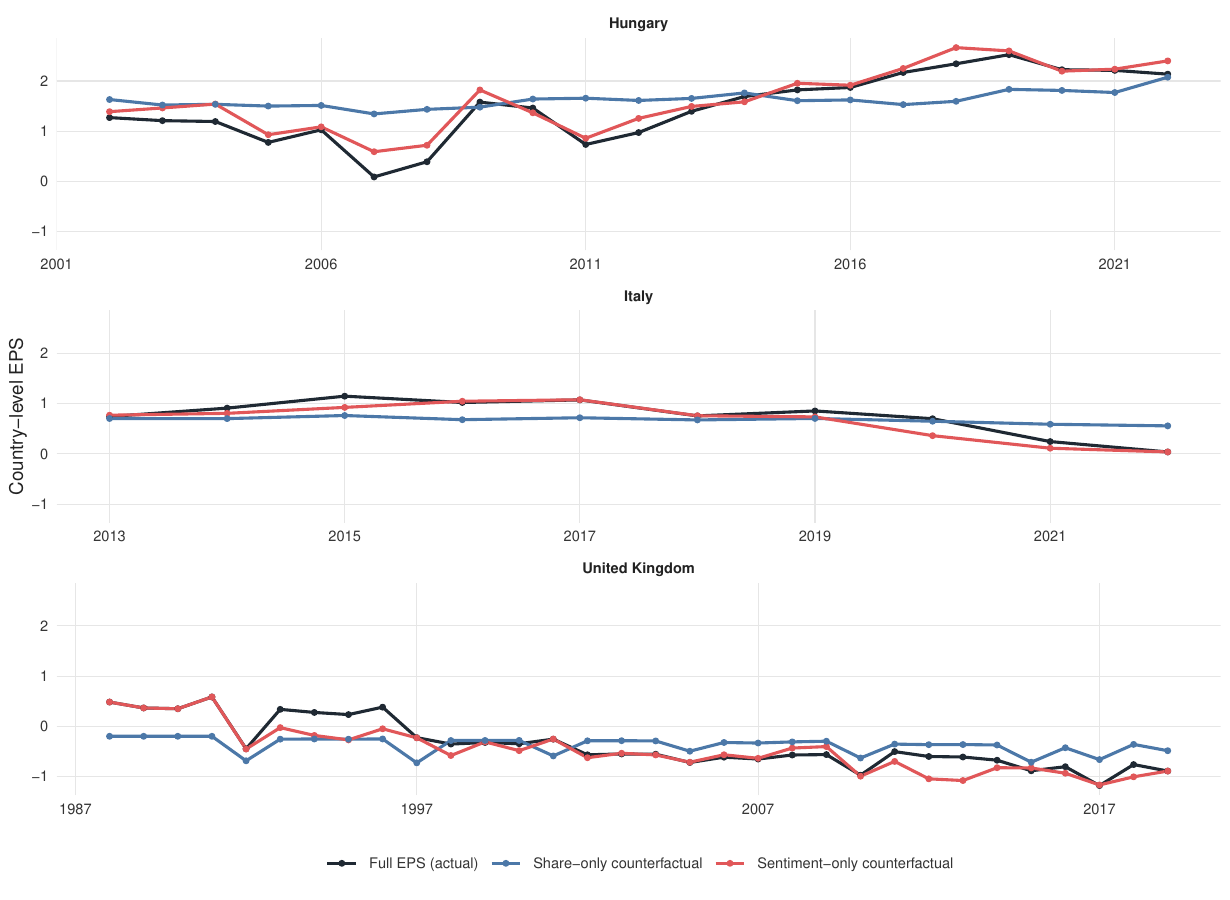}
\caption{Country-level Elite Polarization Score and its share-only and
sentiment-only counterfactuals.\label{fig:i5_decomp}}
\end{figure}

A possible concern is whether movement in the Elite Polarization Score is driven by changes in \emph{which} out-parties are referenced or by changes in \emph{how} those out-parties are evaluated. Figure~\ref{fig:i5_decomp} plots, for each country, the full EPS series alongside two counterfactual series: a \emph{share-only} counterfactual, in which the period's actual reference shares are combined with the long-run pair-baseline mean sentiment; and a \emph{sentiment-only} counterfactual, in which the period's actual mean sentiment per pair is combined with the pair-baseline shares. The sentiment-only series tracks the full series tightly (Pearson $r$ of $0.97$, $0.94$, and $0.92$ for Hungary, Italy, and the United Kingdom, respectively), and reproduces between $83\%$ and $100\%$ of the variance in the full series. The share-only series accounts for at most about $12\%$ of full-series variance. The substantive variation in EPS is therefore driven by shifts in how out-parties are evaluated, not by shifts in the size of out-parties that are referenced.

\end{document}